# Multi-Label Takagi-Sugeno-Kang Fuzzy System

Qiongdan Lou, Zhaohong Deng, *Senior Member, IEEE*, Zhiyong Xiao, Kup-Sze Choi, Shitong Wang

*Abstract*—Multi-label classification can effectively identify the relevant labels of an instance from a given set of labels. However, the modeling of the relationship between the features and the labels is critical to the classification performance. To this end, we propose a new multi-label classification method, called Multi-Label Takagi-Sugeno-Kang Fuzzy System (ML-TSK FS), to improve the classification performance. The structure of ML-TSK FS is designed using fuzzy rules to model the relationship between features and labels. The fuzzy system is trained by integrating fuzzy inference based multi-label correlation learning with multi-label regression loss. The proposed ML-TSK FS is evaluated by experimentally on 12 benchmark multi-label datasets. The results show that the performance of ML-TSK FS is competitive with existing methods in terms of various evaluation metrics, indicating that it is able to model the feature-label relationship effectively using fuzzy inference rules and enhances the classification performance.

*Index Terms*—Multi-label classification, multi-label Takagi-Sugeno-Kang (TSK) fuzzy system, fuzzy inference rules, label correlation learning.

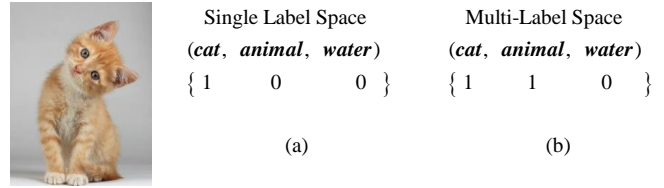

Fig. 1 Example of (a) single label and (b) multi-label classification.

## I. INTRODUCTION

MULTI-LABEL classification is a general form of single label classification. For single label classification, there is only one label assigned to an instance, whereas for multi-label classification, the same instance can be labeled with multiple labels. Fig. 1 shows an example illustrating the difference between single label classification and multi-label classification. In Fig. 1(a), the picture is labeled as belonging to the class of cat, which is a traditional single label classification task. In Fig. 1(b), the picture is labeled as belonging to the class of cat and the class of animal simultaneously, which is a typical multi-label classification task.

According to the strategies adopted, the existing methods of multi-label classification can be divided into two main categories: *problem transformation* and *algorithm adaptation* [1]. They are described as follows:

This work was supported in part by the NSFC under Grants (61772239, 62176105), the Chinese Association for Artificial Intelligence (CAAI)-Huawei MindSpore Open Fund under Grant CAAIXSJLJJ-2021-011A., the Six Talent Peaks Project in Jiangsu Province under Grant XYDXX-056, the Hong Kong Research Grants Council (PolyU 152006/19E), the Hong Kong Innovation and Technology Fund (MRP/015/18) and the Shanghai Municipal Science and Technology Major Project [No.2018SHZDZX01]. (Corresponding author: Zhaohong Deng).

Q. Lou, Z. Deng, Z. Xiao, S. Wang are with the School of Artificial Intelligence and Computer Science, Jiangnan University and Jiangsu Key Laboratory of Digital Design and Software Technology, Wuxi 214122, China. (e-mail: 6171610005@stu.jiangnan.edu.cn; dengzhaohong@jiangnan.edu.cn; zhiyong.xiao@jiangnan.edu.cn; wxwangst@aliyun.com).

K.S. Choi is with the Centre for Smart Health, Hong Kong Polytechnic University. (e-mail: thomasks.choi@polyu.edu.hk).

1) *Problem transformation*: This kind of approaches transforms a multi-label classification problem into other forms of learning problems that can be solved by existing methods, e.g. Binary Relevance (BR) [2] and Classifier Chains (CC) [3], which are two classical problem transformation methods. BR transforms multi-label classification task into multiple binary classification tasks, where each binary classification task can be implemented independently by any binary classifier. Similarly, CC also transforms multi-label classification task into multiple binary classification tasks. Unlike BR, CC regards the classification result of the previous label as a new feature, which is added to the feature space to take part in the classification task of the next label. If there is a logical relationship between the labels, the chain structure of CC can improve the classification performance. In order to improve the learning effect of multi-label model, Zhang et al. proposed multi-label learning with the Label specIfic FeaTures (LIFT) method [4] by first obtaining the specific features of each label by the *k*-means method [5] and then following the same procedure in BR where binary classifiers were employed to classify multiple labels one by one. Besides, Calibrated Label Ranking (CLR) is proposed to transform multi-label problem into multiple label ranking problems [6]. It first converts all labels into label pairs, and then utilizes the existing binary classifiers to solve the binary classification tasks.

2) *Algorithm adaptation*: This kind of approaches is proposed to deal with multi-label problems directly [7-11]. Multi-Label *k*-Nearest Neighbor (ML-*k*NN) [12] is a classical algorithm adaptation method. It calculates the label information of the *k*-nearest neighbors and utilizes the maximized posterior probability to predict the new instance. Zhan et al. proposed the multi-label learning method Label-specIfic FeaTures viA Clustering Ensemble (LIFTACE) [13] by taking label similarity into account to improve the classification performance of the LIFT method. Ensemble of Classifier Chains (ECC) [14] is proposed to improve the stability of the CC method using multiple random label sequences, on which CC is executed individually to obtain the predicted scores of the labels. The predicted labels of an instance are then obtained by averaging the predicted scores. To further improve ECC, Li et al. proposed the Selective Ensemble of Classifier Chains (SECC) [15] to learn the weights of the classifiers in CC by constructing the empirical risk function, so that the computational cost is reduced whilst maintaining the classification performance.

The core task of multi-label classification is to model the relationship between features and labels. On the other hand, it has been revealed in many studies that the classification performance can be improved effectively by utilizing the correlation between different labels [16-19]. Therefore, a promising way for multi-label classification is to design an appropriate model to learn the feature-label relationship and mine the correlation among the labels.

Based on the above analysis, we introduce the fuzzy inference based Takagi-Sugeno-Kang fuzzy system (TSK FS) as the basis model and propose a new multi-label classification method, called Multi-Label TSK FS (ML-TSK FS). The idea of ML-TSK FS is as follows. On the one hand, the rule structure of traditional TSK FS for single label scene is extended for multi-label scene. By the improved TSK fuzzy system structure, the hidden relationship between labels and features can be learned more effectively for multi-label classification. For each label, the corresponding discriminative features can be learned by ML-TSK FS based on its strong nonlinear learning abilities. On the other hand, a label correlation measure mechanism based on fuzzy inference is designed to learn the correlations between the labels. Specifically, each label has its own discriminative features which make significant contribution to the identification of the labels. Moreover, the proposed ML-TSK FS is premised on the assumption that the *correlation between two labels* should be consistent with the *correlation between their discriminative features*. For example, for two labels that are mutually exclusive (or non-intersecting) (e.g. "cat" and "water"), their discriminative features are very different. On the contrary, for two labels that exhibit dependency (or similarity) (e.g. "cat" and "animal"), their discriminative features are partially overlapped. To take advantage of this label correlation information, label correlation measure based on fuzzy inference is integrated into the training of the ML-TSK FS. Based on these mechanisms, the fuzzy inference based label correlation measure and the multi-label regression loss are incorporated into the objective function to train the proposed ML-TSK FS.

The main contributions of this work are summarized as follows:

(1) To model the relationship between features and labels, we modify the traditional TSK FS for multi-label learning by proposing the ML-TSK FS that has two characteristics: (i) the antecedent parameters can be shared by all labels, and (ii) the consequent parameters are designed independently for each label, which can be used to find the discriminative features for different labels.

(2) To leverage the correlation information among labels, label correlation learning based on fuzzy inference is developed, by which correlated discriminative features can be learned effectively for the correlated labels to enhance the classification performance of the ML-TSK FS significantly.

(3) To evaluate the effectiveness of the proposed method, extensive analyses, including the classification performance analysis, correlation analysis, parameter analysis, convergence analysis and statistical analysis, are conducted experimentally on 12 benchmark multi-label datasets.

The rest of the paper is organized as follows. Section II reviews the concepts of multi-label learning and traditional TSK FS. In Section III, the rule structure of traditional single label TSK FS is modified for multi-label scenarios, and the objective function is constructed based on fuzzy inference. The optimization process of the proposed method and the related theoretical analysis are also discussed in Section III. In Section IV, experimental results and analyses are presented. Finally, Section V summarizes the study and discusses the future work.

## II. RELATED WORK

The method proposed in this paper is related to multi-label learning and TSK fuzzy system, which are briefly described in the section.

### A. Problem Definition of Multi-Label Learning

For a multi-label dataset, let $\mathcal{X} = \mathbb{R}^D$ be the feature space with $D$ dimensions, $\mathcal{Y} = \mathbb{R}^L$ be the label space with $L$ dimensions, and $\mathcal{D} = \{(\boldsymbol{x}_i, \boldsymbol{y}_i) | 1 \leq i \leq N\}$ be the training set with $N$ samples. For the $i$th instance $\boldsymbol{x}_i \in \mathbb{R}^{D \times 1}$, denote the ground truth label vector as $\boldsymbol{y}_i = [y_{i1}, y_{i2}, ..., y_{iL}]^T \in \mathbb{R}^{L \times 1}$. If an instance $\boldsymbol{x}_i$ is related to the $j$th label, $y_{ij} = 1$, otherwise $y_{ij} = 0$. The input matrix can be represented as $\boldsymbol{X} = [\boldsymbol{x}_1, \boldsymbol{x}_2, ..., \boldsymbol{x}_N] \in \mathbb{R}^{D \times N}$ and the output matrix as $\boldsymbol{Y} = [\boldsymbol{y}_1, \boldsymbol{y}_2, ..., \boldsymbol{y}_N] \in \mathbb{R}^{L \times N}$. The goal of multi-label learning is to find the mapping function: $f : \mathcal{X} \rightarrow \mathcal{Y}$, which can predict the label vector $\boldsymbol{y}$ for a given new instance $\boldsymbol{x}$.

### B. TSK Fuzzy System

TSK FS is a classical intelligent model based on fuzzy inference rules and fuzzy sets. Attributed to its strong learning ability, TSK FS has been widely used in many fields [20-27]. The core of TSK FS is the fuzzy inference rules. For a classical TSK FS with $K$ fuzzy rules, the $k$th rule can be described as follows:

$$\begin{aligned} &\text{IF}: x_1 \text{ is } A_1^k \wedge x_2 \text{ is } A_2^k \wedge \cdots \wedge x_D \text{ is } A_D^k, \\ &\text{THEN}: f^k(\boldsymbol{x}) = p_0^k + p_1^k x_1 + \cdots + p_D^k x_D, \\ &\qquad k = 1, 2, ..., K. \end{aligned} \quad (1)$$

where $D$ is the dimension of the feature space and $K$ is the number of fuzzy rules; $x_j$ ($j = 1, 2, ..., D$) is the $j$th feature of the input vector $\boldsymbol{x}$; $A_j^k$ is the antecedent fuzzy set corresponding to the $j$th feature of the input vector $\boldsymbol{x}$ in the $k$th rule; $\wedge$ is a fuzzy conjunction operator; $f^k(\boldsymbol{x})$ is the output of the input vector $\boldsymbol{x}$ under the $k$th fuzzy inference rule; and $p_j^k$ ($j = 0, 1, 2, ..., D$) is the consequent parameter of the $k$th rule.

In TSK FS, membership functions are needed to describe the antecedent fuzzy set $A_j^k$ in (1). Different membership functions can be used for different scenarios. In this paper, Gaussian function is adopted for its wide application in various fields [28]. It is defined as:

$$\mu_{A_i^k}(x_i) = \exp(-(x_i - c_i^k)^2 / 2(\delta_i^k)^2) \qquad (2)$$

where the parameters $c_i^k$ and $\delta_i^k$ can be estimated using different methods. In the absence of domain knowledge, we usually use data-driven methods to estimate $c_i^k$ and $\delta_i^k$. Here, the commonly used Fuzzy $C$-Means (FCM) clustering algorithm [29] is employed to obtain the two parameters of the Gaussian function, which are given by

$$c_i^k = \sum_{j=1}^{N} u_{jk} x_{ji} \Big/ \sum_{j=1}^{N} u_{jk} \tag{3}$$

$$\delta_i^k = h \cdot \sum_{j=1}^{N} u_{jk} (x_{ji} - c_i^k)^2 \Big/ \sum_{j=1}^{N} u_{jk} \tag{4}$$

where $u_{jk}$ represents the membership value of the $j$th instance $x_j$ on the $k$th rule. The value of $u_{jk}$ can be obtained by FCM. $h$ is a hyperparameter to adjust the antecedent parameter $\delta_i^k$.

Corresponding to the $k$th fuzzy set $A_i^k$, if the membership value of the $i$th feature $x_i$ is $\mu_{A_i^k}(x_i)$, the firing strength of instance $x$ on the $k$th rule can be defined as:

$$\mu^k(x) = \prod_{i=1}^{D} \mu_{A_i^k}(x_i) \tag{5}$$

$$\tilde{\mu}^k(x) = \mu^k(x) \Big/ \sum_{k'=1}^{K} \mu^{k'}(x), \tag{6}$$

where (6) is the standardized form of (5). Finally, the output of instance $x$ in TSK FS can be expressed as

$$y = f_{TSK-FS}(x) = \sum_{k=1}^{K} \tilde{\mu}^k(x) f^k(x), \tag{7}$$

which can be further expressed as a linear model of the high-dimensional feature space mapped by fuzzy rules, i.e.,

$$y = f_{TSK-FS}(x) = x_g^T p_g \tag{8}$$

where,

$$x_e = [1, x^T]^T \in \mathbb{R}^{(D+1) \times 1} \tag{9}$$

$$\tilde{x}^k = \tilde{\mu}^k(x) x_e \in \mathbb{R}^{(D+1) \times 1} \tag{10}$$

$$x_g = [(\tilde{x}^1)^T, (\tilde{x}^2)^T, ..., (\tilde{x}^K)^T]^T \in \mathbb{R}^{K(D+1) \times 1} \tag{11}$$

$$p^k = [p_0^k, p_1^k, ..., p_D^k]^T \in \mathbb{R}^{(D+1) \times 1} \tag{12}$$

$$p_g = [(p^1)^T, (p^2)^T, ..., (p^K)^T]^T \in \mathbb{R}^{K(D+1) \times 1} \tag{13}$$

$x_g$ is a vector obtained for the instance $x$ by fuzzy rule mapping. $p_g$ is a vector containing all the consequent parameters of TSK FS.

## III. MULTI-LABEL TAKAGI-SUGENO-KANG FUZZY SYSTEM

### A. Framework

In this paper, the ML-TSK FS is proposed to utilize the label correlation information and model the inference relationship between features and labels. The framework is shown in Fig. 2, which involves two core components: (1) the design of rule structure, and (2) the construction of objective function. The goal of the former is to develop an appropriate structure of fuzzy rules to model the feature-label relationship more conveniently; whereas the aim of the latter is to make the learning of the ML-TSK FS more effective by integrating fuzzy inference based label correlation learning with multi-label regression loss, so as to improve the performance of the traditional TSK FS learning method for multi-label learning scene.

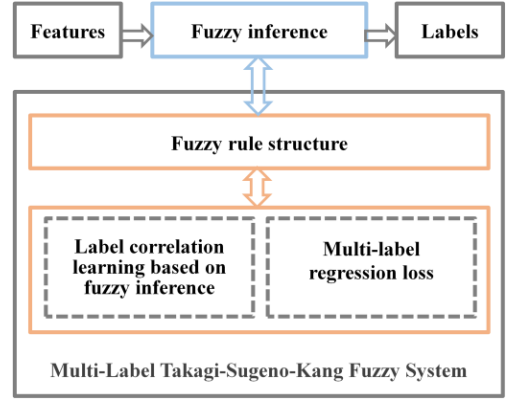

Fig. 2 The framework of the proposed method ML-TSK FS.

The details of the proposed ML-TSK FS are discussed in the following five sections: the design of the rule structure of ML-TSK FS in Section III-B, the construction of the objective function of ML-TSK FS in Section III-C, the optimization of ML-TSK FS in Section III-D, algorithm description in Section III-E, and the analysis of computational complexity in Section III-F.

### B. Rule Structure of ML-TSK FS

TSK FS is a classical fuzzy system model that has been successfully applied to many fields, attributed to its simplicity and effectiveness in modeling fuzzy inference [30-35]. To extend it for multi-label scenarios, we propose the ML-TSK FS by modifying the rule structure of the traditional TSK FS. In multi-label learning, ML-TSK FS generates only one set of antecedent parameters and one consequent parameter matrix. Specifically, for ML-TSK FS with $K$ rules, the $k$th rule can be expressed as:

IF: $x_1$ is $A_1^k \wedge x_2$ is $A_2^k \wedge \cdots \wedge x_D$ is $A_D^k$,

THEN: $f^k(x) = \begin{bmatrix} f_1^k(x) \\ \vdots \\ f_l^k(x) \\ \vdots \\ f_L^k(x) \end{bmatrix} = \begin{bmatrix} p_{10}^k + p_{11}^k x_1 + \cdots + p_{1D}^k x_D \\ \vdots \\ p_{l0}^k + p_{l1}^k x_1 + \cdots + p_{lD}^k x_D \\ \vdots \\ p_{L0}^k + p_{L1}^k x_1 + \cdots + p_{LD}^k x_D \end{bmatrix} \begin{matrix} (\text{label 1}) \\ \vdots \\ (\text{label } l) \\ \vdots \\ (\text{label } L) \end{matrix}$

$1 \le k \le K$ and $1 \le l \le L$

(14)

where $D$, $K$, and $L$ are the number of features, the number of fuzzy rules and the number of labels respectively. For an instance $x$, $A_j^k$ represents the antecedent fuzzy set of the $j$th ($1 \le j \le D$) feature in the $k$th rule, and $f_l^k(x)$ is the output of $x$ for the $l$th label in the $k$th rule.

Let $p_{g,l} = [(p_l^1)^T, (p_l^2)^T, ..., (p_l^K)^T]^T \in \mathbb{R}^{K(D+1) \times 1}$ be the consequent parameter vector corresponding to the $l$th label ($1 \le l \le L$), and $p_l^k = [p_{l0}^k, p_{l1}^k, ..., p_{lD}^k]^T \in \mathbb{R}^{(D+1) \times 1}$ ($1 \le l \le L$) be the consequent parameter vector corresponding to the $l$th label under the $k$th rule. Hence, the complete consequent parameter matrix of the ML-TSK FS can be defined as:

$$P = [p_{g,1}, p_{g,2}, ..., p_{g,L}] \in \mathbb{R}^{K(D+1) \times L} \quad (15)$$

The antecedent set $A_j^k$ is represented by the membership function of the $j$th feature in the $k$th rule. In this paper, Gaussian function is used as the membership function, and the FCM clustering algorithm is used to estimate the corresponding parameters. For the $k$th rule, the standardized membership $\tilde{\mu}^k(x)$ of instance $x$ can be obtained from (2)-(6).

Finally, the output of instance $x$ in the ML-TSK FS can be expressed as:

$$y_{ML-TSKFS} = f_{ML-TSKFS}(x) = [f_1(x), f_2(x), ..., f_L(x)]^T$$
$$= [\sum_{k=1}^{K} \tilde{\mu}^k(x) f_1^k(x), \sum_{k=1}^{K} \tilde{\mu}^k(x) f_2^k(x), ..., \sum_{k=1}^{K} \tilde{\mu}^k(x) f_L^k(x)]^T \quad (16)$$

(16) can be re-expressed as a linear model of high-dimensional features mapped by the fuzzy rules:

$$y_{ML-TSKFS} = f_{ML-TSKFS}(x)$$
$$= \begin{bmatrix} p_{10}^1 \cdot \tilde{\mu}^1(x) + \cdots + p_{1D}^1 \cdot (\tilde{\mu}^1(x) x_D) + \cdots + p_{10}^K \cdot \tilde{\mu}^K(x) + \cdots + p_{1D}^K \cdot (\tilde{\mu}^K(x) x_D) \\ p_{20}^1 \cdot \tilde{\mu}^1(x) + \cdots + p_{2D}^1 \cdot (\tilde{\mu}^1(x) x_D) + \cdots + p_{20}^K \cdot \tilde{\mu}^K(x) + \cdots + p_{2D}^K \cdot (\tilde{\mu}^K(x) x_D) \\ \vdots \\ p_{L0}^1 \cdot \tilde{\mu}^1(x) + \cdots + p_{LD}^1 \cdot (\tilde{\mu}^1(x) x_D) + \cdots + p_{L0}^K \cdot \tilde{\mu}^K(x) + \cdots + p_{LD}^K \cdot (\tilde{\mu}^K(x) x_D) \end{bmatrix}$$
$$= [p_{g,1}^T x_g, p_{g,2}^T x_g, ..., p_{g,L}^T x_g]^T$$
$$= P^T x_g \quad (17)$$

where $x_g$ is a vector obtained for the instance $x$ by fuzzy rule mapping and the process is detailed in (2)-(11). $P \in \mathbb{R}^{K(D+1) \times L}$ is the consequent parameter matrix of the ML-TSK FS, and its definition is shown in (15).

Compared with traditional TSK FS, ML-TSK FS has the following advantages. First, all the labels share a common set of antecedent parameters, which reduces the burden of multi-label learning. Second, the consequent parameters of the ML-TSK FS are in the form of the matrix $P$ that each column corresponds to a unique label. The matrix $P$ has two functions. First, it models the mapping relationship between features and labels by multi-label regression. Second, the parameter of the inference rules $P$ is used in the label correlation learning to leverage the label correlation information. In addition, $P$ can be used to identify the discriminative feature in each rule for different labels. Therefore, the main training task of ML-TSK FS is to learn the consequent parameter matrix $P$. The learning process of $P$ is detailed in the next section.

### C. Objective Function of ML-TSK FS

The learning procedure of ML-TSK FS contains two main parts: modeling the fuzzy inference between features and labels, and utilizing the label correlation information. For the first part, we exploit multi-label regression loss for parameter optimization. For the second, we introduce label correlation learning based on fuzzy inference. Specifically, each label has its own discriminative features that are indicated by the corresponding column in $P$ (i.e., consequent parameter vector). In addition, we premise on the reasonable assumption that the correlation between two labels is consistent with the correlation between their discriminative features, which has been elaborated in Section I. Therefore, the label information can be utilized by considering the correlation between any two labels, and their discriminative features.

Based on the above analysis, we take the consequent parameter matrix $P$ as the solution variable and define the general form of the objective function for ML-TSK FS as follows.

$$P^* = \arg\min_P \mathcal{L}_{re}(P) + \alpha \cdot \mathcal{L}_{corr}(P) \quad (18)$$

In (18), $\mathcal{L}_{re}(P)$ represents the multi-label regression loss and $\mathcal{L}_{corr}(P)$ represents the label correlation learning based on fuzzy inference. Here, $\alpha$ is a nonnegative hyperparameter that balances the effect of the two terms.

#### 1) Multi-Label Regression Loss

Given a training set $X = [x_1, x_2, ..., x_N] \in \mathbb{R}^{D \times N}$ that can be mapped to a new feature set $G = [x_{g1}, x_{g2}, ..., x_{gN}] \in \mathbb{R}^{K(D+1) \times N}$ through the fuzzy rules in the ML-TSK FS as detailed in (2)-(11). $P$ contains $L$ consequent parameter vectors (column vectors), and every vector corresponds to a specific label. In the new feature space, $p_{g,l} = [(p_l^1)^T, (p_l^2)^T, ..., (p_l^K)^T]^T \in \mathbb{R}^{K(D+1) \times 1}$ corresponds to the $l$th label. In general, TSK FS adopts linear regression to deal with single label scenes, that is,

$$\mathcal{L}_{re}(p_{g,l}) = \frac{1}{2} \left\| p_{g,l}^T G - \tilde{y}_l \right\|_2^2 \quad (19)$$

where $\tilde{y}_l = [y_{1l}, y_{2l}, ..., y_{nl}, ..., y_{Nl}]$ $(1 \leq l \leq L)$, $y_{nl}$ is the $l$th label of the $n$th sample, and $N$ is the number of training samples. That is, $\tilde{y}_l$ is a vector composed of the $l$th row elements in the training label matrix $Y \in \mathbb{R}^{L \times N}$.

Correspondingly, for ML-TSK FS, the following regression loss function can be constructed.

$$\mathcal{L}_{re}(P) = \frac{1}{2} \left\| P^T G - Y \right\|_F^2 \quad (20)$$

To improve the robustness of the fuzzy system, (20) can be further expressed as:

$$\mathcal{L}_{re}(P) = \frac{1}{2} \left\| P^T G - Y \right\|_F^2 + \beta \left\| P \right\|_1 \quad (21)$$

where $\left\| P \right\|_1$ is the regularization item, and $\beta$ is a positive hyperparameter to balance the regression loss and the complexity of the model. In this paper, we adopt the L1 norm which is widely used in machine learning [36-40]. Compared with the steadily differentiable L2 norm, the L1 norm in (21) has the following advantages:

(1) With the L1 norm, it is easier to obtain a sparse solution for the consequent parameters of the fuzzy system $p_{g,l}$ $(1 \leq l \leq L)$ to predict the $l$th label $\tilde{y}_l$. Since most of the elements in $p_{g,l}$ tend to 0, the resulting multi-label TSK FS would be more concise. Hence, L1 norm can effectively reduce the model complexity and computational complexity of the multi-label model.

(2) For each pair $(p_{g,l}, \tilde{y}_l)$ $(1 \leq l \leq L)$, the feature subset of the input $x_{gi}$ corresponding to the non-zero parameters in $p_{g,l}$ are more discriminative for the label $\tilde{y}_l$. That is, discriminative features for different labels can be obtained using the L1 norm.

(3) Since the optimized $p_{g,l}$ $(1 \leq l \leq L)$ obtained with the L1 norm is sparse, the important features in different rules can be readily identified while the minor ones can be ignored, yielding simple and clear inference relationship between features and the $l$th label $\tilde{y}_l$.

*2) Label Correlation Learning Based on Fuzzy Inference*

To utilize the label correlation information, label correlation learning is introduced based on fuzzy inference. Correlation information between labels can be described in different ways, such as conditional dependence network and covariance matrix [16, 18], and the Pearson correlation coefficient [41] is used here.

In this paper, the label correlation information is extracted for the proposed ML-TSK FS based on the following arguments: 1) the $i$th consequent parameter vector $p_{g,i} \in \mathbb{R}^{K(D+1) \times 1}$ in $P \in \mathbb{R}^{K(D+1) \times L}$ can be used to find the discriminative features for the $i$th label $\tilde{y}_i \in \mathbb{R}^{1 \times N} (1 \leq i \leq L)$ in $Y \in \mathbb{R}^{L \times N}$; and 2) the correlation between two labels is consistent with the correlation between their discriminative features. Here, the label matrix $Y$ is known, and $P$ is set as the consequent parameter matrix to be solved. The correlation between labels can be quantified by $Y$ using the Pearson correlation function as the correlation measurement technique. The quantified correlation between $\tilde{y}_i$ $(1 \leq i \leq L)$ and $\tilde{y}_j$ $(1 \leq j \leq L)$ can then be used to train the correlation between $p_{g,i}$ and $p_{g,j}$. Since the consequent parameter vectors determine the discriminative features for each label, the above mechanism facilitates the extraction of discriminative features for label prediction, and the multi-label classification performance would be improved accordingly, i.e., the correlation learning can benefit multi-label classification effectively.

Thus, the label correlation learning based on fuzzy inference is constructed as follows:

$$\mathcal{L}_{corr}(p_{g,i}, p_{g,j}) = \frac{1}{2} r_{ij} p_{g,i}^T p_{g,j}, \quad (22)$$

where $r_{ij} = 1 - c_{ij}$ is used to quantify the correlation between labels $\tilde{y}_i$ $(1 \leq i \leq L)$ and $\tilde{y}_j$ $(1 \leq j \leq L)$, and $c_{ij}$ is the correlation coefficient between labels $\tilde{y}_i$ and $\tilde{y}_j$. $c_{ij}$ is calculated using the Pearson correlation function in this paper. The stronger the correlation between $\tilde{y}_i$ and $\tilde{y}_j$, the larger the value of $c_{ij}$ and the smaller the value of $r_{ij}$, and vice versa. For the complete multi-label space, the function of correlation learning is defined as follows:

$$\mathcal{L}_{corr}(P) = \frac{1}{2} \sum_{i=1}^{L} \sum_{j=1}^{L} r_{ij} p_{g,i}^T p_{g,j} \\ = \frac{1}{2} Tr(RP^T P) \quad (23)$$

where $P \in \mathbb{R}^{K(D+1) \times L}$ is defined in (15), and $R = [r_{ij}] \in \mathbb{R}^{L \times L}$. (23) can be used to guide the optimization of the consequent parameter matrix $P$.

*3) The Overall Objective Function*

By combining (18), (21) and (23), the final objective function can be expressed as:

$$P^* = \arg\min_{P} \frac{1}{2} \|P^T G - Y\|_F^2 + \beta \|P\|_1 + \frac{\alpha}{2} Tr(RP^T P) \quad (24)$$

For a new test instance $x'$, the prediction output of the ML-TSK FS can be expressed as:

$$y'_{ML-TSKFS} = f_{ML-TSKFS}(x') = (P^*)^T x'_g, \quad (25)$$

where $y'_{ML-TSKFS} = [y'_{ML-TSKFS,1}, y'_{ML-TSKFS,2}, ..., y'_{ML-TSKFS,L}]^T$ and $y'_{ML-TSKFS} \in \mathbb{R}^{L \times 1}$ is the output predicted by the ML-TSK FS, and $x'_g$ is mapped by $x'$ through the fuzzy rules. Further, the threshold function $\varphi(\cdot)$ is introduced to improve the flexibility of the model. That is, the predicted label vector $y' = [y'_1, y'_2, ..., y'_L]^T$ can be obtained by the following threshold function:

$$y'_l = \varphi_\tau(y'_{ML-TSKFS,l}) = \begin{cases} 1, & \text{if } y'_{ML-TSKFS,l} > \tau \\ 0, & \text{otherwise} \end{cases} \quad (1 \leq l \leq L), \quad (26)$$

where $\tau$ is the adjustable threshold. In fact, the value of $\tau$ can be optimized by cross-validation. In this paper, we fix it to the commonly adopted value of 0.5.

*D. Objective Function Optimization*

Since the objective function of ML-TSK FS contains the L1 norm of $P$, which is not differentiable with respect to $P$, we cannot directly obtain the gradients on $P$ for optimization. Efficient optimization techniques have been developed to solve this common problem in L1 norm-based methods. In the study, the commonly used techniques, the Proximal Gradient Descent (PGD) [42], is utilized to solve for the parameter $P$ of the proposed ML-TSK FS. The optimization process is described below.

For the proposed ML-TSK FS, the complete objective function is given by

$$\begin{aligned} P^* &= \arg\min_{P} \frac{1}{2} \|P^T G - Y\|_F^2 + \beta \|P\|_1 + \frac{\alpha}{2} Tr(RP^T P) \\ &= \arg\min_{P} f(P) + \beta \|P\|_1 \end{aligned} \quad (27)$$

with

$$f(P) = \frac{1}{2} \|P^T G - Y\|_F^2 + \frac{\alpha}{2} Tr(RP^T P). \quad (28)$$

Since both $f(P)$ and the L1 norm $\|P\|_1$ are convex, and $\beta > 0$ as defined in (21), the optimization problem in (27) is also convex (the proof is shown in the section *Supplementary Materials: Part A*). Besides, the function in (28) is convex and smooth, with a Lipschitz constant (the proof is shown in the section *Supplementary Materials: Part B*). Under these conditions, there exists a constant $L_f > 0$ such that

$$\|\nabla f(P') - \nabla f(P'')\| \leq L_f \|P' - P''\| \quad (\forall P', P''). \quad (29)$$

Then, (27) can be solved in an iterative manner. For the $t$th iteration, given the fixed point $P^{(t)}$, $f(P)$ can be approximated using second-order Taylor expansion

$$\hat{f}(\boldsymbol{P}) \simeq f(\boldsymbol{P}^{(t)}) + \langle \nabla f(\boldsymbol{P}^{(t)}), \boldsymbol{P} - \boldsymbol{P}^{(t)} \rangle + \frac{L_f}{2} \|\boldsymbol{P} - \boldsymbol{P}^{(t)}\|_F^2$$
$$= \frac{L_f}{2} \left\| \boldsymbol{P} - (\boldsymbol{P}^{(t)} - \frac{1}{L_f} \nabla f(\boldsymbol{P}^{(t)})) \right\|_F^2 + const \quad (30)$$

where $const$ is a constant which is independent of $\boldsymbol{P}$. Therefore, for the $(t+1)$th iteration, (27) can be approximated by

$$\boldsymbol{P}_{t+1} = \arg\min_{\boldsymbol{P}} \ \hat{f}(\boldsymbol{P}) + \beta \|\boldsymbol{P}\|_1$$
$$= \arg\min_{\boldsymbol{P}} \ \frac{L_f}{2} \left\| \boldsymbol{P} - (\boldsymbol{P}^{(t)} - \frac{1}{L_f} \nabla f(\boldsymbol{P}^{(t)})) \right\|_F^2 + \beta \|\boldsymbol{P}\|_1, \quad (31)$$
$$= \arg\min_{\boldsymbol{P}} \ \frac{1}{2} \|\boldsymbol{P} - \boldsymbol{Z}^{(t)}\|_F^2 + \frac{\beta}{L_f} \|\boldsymbol{P}\|_1$$

where $\boldsymbol{Z}^{(t)} = \boldsymbol{P}^{(t)} - \nabla f(\boldsymbol{P}^{(t)})/L_f$. Then, (31) can be solved by the following update rule (the details are shown in the section *Supplementary Materials: Part C*):

$$\boldsymbol{P}_{t+1} = S_{\beta/L_f}[\boldsymbol{Z}^{(t)}], \quad (32)$$

where $S_{\beta/L_f}[\boldsymbol{Z}^{(t)}]$ is the soft threshold function, which is defined by $\boldsymbol{Z}^{(t)} = [z_{ij}]$ and $\beta/L_f$ as follows:

$$\left(S_{\beta/L_f}[\boldsymbol{Z}^{(t)}]\right)_{ij} = \begin{cases} z_{ij} - \beta/L_f, & if \ z_{ij} > \beta/L_f \\ z_{ij} + \beta/L_f, & if \ z_{ij} < -\beta/L_f \\ 0, & otherwise \end{cases} \quad (33)$$

In addition, to obtain the optimal solution for (27) more effectively, we first obtain the solution of (20) and take it as the start value (i.e., the initial value of $\boldsymbol{P}$) of the subsequent iterations in the learning procedure for solving (27). The process is as follows:

The derivative of (20) with respect to $\boldsymbol{P}$ is
$$\nabla \mathcal{L}_{re}(\boldsymbol{P}) = \boldsymbol{G}\boldsymbol{G}^T \boldsymbol{P} - \boldsymbol{G}\boldsymbol{Y}^T \quad (34)$$

Let $\nabla \mathcal{L}_{re}(\boldsymbol{P}) = 0$, we get
$$\boldsymbol{P} = (\boldsymbol{G}\boldsymbol{G}^T)^{-1} \boldsymbol{G}\boldsymbol{Y}^T \quad (35)$$

Therefore, we set the initial value of $\boldsymbol{P}$ as
$$\boldsymbol{P}_0 = (\boldsymbol{G}\boldsymbol{G}^T + \gamma \boldsymbol{I})^{-1} \boldsymbol{G}\boldsymbol{Y}^T \quad (36)$$

where $\gamma$ is a hyperparameter, and the introduction of $\gamma \boldsymbol{I}$ can further ensure the stability of the solution.

To improve the convergence speed of ML-TSK FS, we reset the fixed point $\boldsymbol{P}^{(t)}$ in (31) in each iteration by updating it to $\boldsymbol{P}^{(t)} = \boldsymbol{P}_t + ((b_{t-1}-1)/b_t)(\boldsymbol{P}_t - \boldsymbol{P}_{t-1})$, where the sequence $(b_t)$ satisfies $b_{t+1}^2 - b_{t+1} \leq b_t^2$, and $\boldsymbol{P}_t$ is the result of the *t*th iteration [43].

### E. Algorithm Description

Based on the above analysis, the training procedure of the ML-TSK FS is given in Algorithm I.

### F. Computational Complexity

The computational complexity of ML-TSK FS in Algorithm I is analyzed step by step as follows. For step 1, the computational complexity of learning the antecedent parameters is $O(2DNK)$. For step 2, by using (2)-(6) and (9)-(11), the complexity of transforming $\boldsymbol{X}$ into $\boldsymbol{G}$ is $O(2NKD+2NK)$. For step 3, the complexity of initializing the matrix $\boldsymbol{P}_0$ and the matrix $\boldsymbol{P}_1$ is $O(K^2N(D+1)^2 + KNL(D+1))$. The complexity of step 5 is $O(1)$. In step 6, the complexity of evaluating $\nabla f(\boldsymbol{P}^{(0)})$ is $O(TK^2N(D+1)^2 + TK^2L(D+1)^2 + TKNL(D+1) + TKL^2(D+1))$. The complexity of step 7 is $O(1)$. Among the steps discussed above, steps 3 and 6 are the determining steps that dominate the computational complexity. Let $a = \max(L, T, K)$, $b = \max(N, K(D+1))$, where $a \ll b$ in general, the overall complexity of the whole algorithm can be expressed as $O(ab^2(a+2b))$.

---

**Algorithm I** ML-TSK FS

**Input**: Training data matrix $\boldsymbol{X} \in \mathbb{R}^{D \times N}$, label matrix $\boldsymbol{Y} \in \mathbb{R}^{L \times N}$, number of fuzzy rules $K$, trade-off parameters $h$, $\alpha$, $\beta$ and $\gamma$.

**Procedure ML-TSK FS:**

1: Learn antecedent parameters $c_i^k$ and $\delta_i^k$ using FCM and (3)-(4);
2: Transform the input matrix $\boldsymbol{X}$ to the fuzzy matrix $\boldsymbol{G}$ using (2)-(6) and (9)-(11);
3: $b_0, b_1 \leftarrow 1$, $\boldsymbol{P}_0, \boldsymbol{P}_1 \leftarrow (\boldsymbol{G}\boldsymbol{G}^T + \gamma \boldsymbol{I})^{-1}\boldsymbol{G}\boldsymbol{Y}^T$, and $t \leftarrow 1$;
4: **While** *not converged* **do**
5:     $\boldsymbol{P}^{(t)} \leftarrow \boldsymbol{P}_t + ((b_{t-1}-1)/b_t)(\boldsymbol{P}_t - \boldsymbol{P}_{t-1})$;
6:     $\boldsymbol{Z}^{(t)} \leftarrow \boldsymbol{P}^{(t)} - (1/L_f)\nabla f(\boldsymbol{P}^{(t)})$;
7:     $\boldsymbol{P}_{t+1} \leftarrow S_{\beta/L_f}[\boldsymbol{Z}^{(t)}]$;
8:     $b_{t+1} \leftarrow (1+\sqrt{4b_t^2+1})/2$;
9:     $t \leftarrow t+1$;
10:    Check the convergence conditions;
11: **End**

**Output**: $\boldsymbol{P}$.

---

TABLE I
DESCRIPTIONS OF DATASETS

| Dataset | #Instance | #Feature | #Label |
|---|---|---|---|
| Bibtex | 7395 | 1836 | 159 |
| Birds | 645 | 260 | 19 |
| CAL500 | 502 | 68 | 174 |
| Corel16k1 | 13766 | 500 | 153 |
| Emotions | 593 | 72 | 6 |
| Flags | 194 | 19 | 7 |
| Image | 600 | 294 | 5 |
| Mirflickr | 25000 | 1000 | 38 |
| Rcv1s1 | 6000 | 944 | 101 |
| Rcv1s2 | 6000 | 944 | 101 |
| Scene | 2407 | 294 | 6 |
| Yeast | 2417 | 103 | 14 |

## IV. EXPERIMENTAL ANALYSIS

We have conducted extensive experimental analysis to fully evaluate the proposed ML-TSK FS. The evaluation includes the analysis of the classification result, label correlation, parameter settings, convergence, and the statistical performance. The datasets, evaluation metrics, and settings of the experiments are described as follows.

### A. Datasets

The performance of ML-TSK FS is evaluated using 12 benchmark multi-label datasets. The Mirflickr dataset is obtained from [44]. The Image dataset can be downloaded from

Github[1]. Other datasets can be obtained from MULAN[2]. Table I gives the description of the datasets, where #Instance, #Feature, and #Label denote the instance number, the feature dimension, and the label space dimension respectively.

### B. Evaluation Metrics

In the test set $X_t = [\boldsymbol{x}_{t,1}, \boldsymbol{x}_{t,2},...,\boldsymbol{x}_{t,N_t}] \in \mathbb{R}^{D \times N_t}$ with $N_t$ samples, $f(\boldsymbol{x}_{t,i}, l)$ represents the continuous output of the prediction model on the $l$th label. The ranking function $rank_f(\boldsymbol{x}_{t,i}, l)$ can be defined according to $f(\boldsymbol{x}_{t,i}, l)$. If $f(\boldsymbol{x}_{t,i}, l') > f(\boldsymbol{x}_{t,i}, l)$, then $rank_f(\boldsymbol{x}_{t,i}, l') < rank_f(\boldsymbol{x}_{t,i}, l)$. $\mathcal{Y} = \{0,1\}^L$ denotes the label space, $\hat{\boldsymbol{y}}_{t,i}$ is the prediction result of $\boldsymbol{x}_{t,i} (1 \leq i \leq N_t)$, and $\boldsymbol{y}_{t,i}$ is the true label vector of $\boldsymbol{x}_{t,i}$. $L_{\boldsymbol{x}_{t,i}}$ is the set of labels associated with $\boldsymbol{x}_{t,i}$, and $\bar{L}_{\boldsymbol{x}_{t,i}}$ is the complement of $L_{\boldsymbol{x}_{t,i}}$. In order to measure the performance of multi-label classifier from different perspectives, five evaluation metrics for multi-label learning are adopted [10]:

(1) Average Precision (AP): It evaluates the ratio of the related label ranking before a certain label $l$. The larger the value of AP, the better the classification performance.

$$\text{AP} = \frac{1}{N_t} \sum_{i=1}^{N_t} \frac{1}{|L_{\boldsymbol{x}_{t,i}}|} \sum_{l \in L_{\boldsymbol{x}_{t,i}}} \frac{|\{l' \in L_{\boldsymbol{x}_{t,i}} | f(\boldsymbol{x}_{t,i}, l') \geq f(\boldsymbol{x}_{t,i}, l)\}|}{rank_f(\boldsymbol{x}_{t,i}, l)}$$

(2) Hamming Loss (HL): It evaluates the proportion of labels that are predicted incorrectly. The smaller the value of HL, the better the classification performance.

$$\text{HL} = \frac{1}{N_t} \sum_{i=1}^{N_t} |\hat{\boldsymbol{y}}_{t,i} \oplus \boldsymbol{y}_{t,i}| / L$$

where $\oplus$ represents XOR operation.

(3) One Error (OE): It evaluates the proportion of labels that have the largest predicted values but the prediction is wrong. The smaller the value of OE, the better the classification performance.

$$\text{OE} = \frac{1}{N_t} \sum_{i=1}^{N_t} \pi(\boldsymbol{x}_{t,i}), \quad \pi(\boldsymbol{x}_{t,i}) = \begin{cases} 1, & \text{if } \arg\max_l f(\boldsymbol{x}_{t,i}, l) \notin L_{\boldsymbol{x}_{t,i}} \\ 0, & \text{otherwise} \end{cases}$$

(4) Ranking Loss (RL): It evaluates the scale of the label pairs that are ranked incorrectly. The smaller the value of RL, the better the classification performance.

$$\text{RL} = \frac{1}{N_t} \sum_{i=1}^{N_t} \frac{|\{(l,l') | f(\boldsymbol{x}_{t,i}, l) \leq f(\boldsymbol{x}_{t,i}, l'), (l,l') \in L_{\boldsymbol{x}_{t,i}} \times \bar{L}_{\boldsymbol{x}_{t,i}}\}|}{|L_{\boldsymbol{x}_{t,i}}||\bar{L}_{\boldsymbol{x}_{t,i}}|}$$

(5) Coverage (CV): It evaluates the average number of times that all related labels of an instance are found. The smaller the value of CV, the better the classification performance.

$$\text{CV} = \frac{1}{N_t} \sum_{i=1}^{N_t} \max_{l \in L_{\boldsymbol{x}_{t,i}}} rank_f(\boldsymbol{x}_{t,i}, l) - 1$$

### C. Experiment Settings

The proposed ML-TSK FS is compared with six algorithm adaptation methods and two problem transformation methods. The six algorithm adaptation methods are ML-$k$NN [12], multi-label classification with Meta-Label-Specific Features (MLSF) [10], Hybrid Noise-Oriented Multi-label Learning (HNOML) [45], Canonical Correlated AutoEncoder (C2AE) [44], Backpropagation for Multi-Label Learning (BP-MLL) [46], and Joint Binary Neural Network (JBNN) [47]. The last three methods, C2AE, BP-MLL and JBNN, are deep learning-based methods. The two problem transformation methods are BR and CC, which have been mentioned in Section I. A brief description of these methods is given as follows.

ML-$k$NN: The method utilizes the label information of the $k$-nearest neighbors to predict the new instance.

MLSF: The method adopts meta-label learning and specific feature selection approach to fully consider the correlation between the labels.

HNOML: The method utilizes bi-sparsity regularization and label enrichment to deal with label noise and feature noise, which can improve the robustness of the method.

C2AE: The method adopts deep canonical correlation analysis and the autoencoder structure, which are determined by feature mapping, encoding function and decoding function.

BP-MLL: The method considers correlation learning by assuming that the rank of related labels is higher than that of unrelated labels.

JBNN: The method exploits multiple logistic functions instead of a softmax function in the network for different labels. In addition, it utilizes a joint binary cross entropy loss to capture the label correlation.

BR: The method uses the $\varepsilon$-Insensitive Learning by Solving a System of Linear Inequalities ($\varepsilon$LSSLI) [48] as the binary classification method. The $\varepsilon$LSSLI improves the generalization ability of fuzzy classification by introducing $\varepsilon$-insensitive learning.

CC: The method decomposes multi-label classification task into multiple binary classification tasks, and each binary classification task is completed by $\varepsilon$LSSLI with different parameter settings.

Five-fold cross-validation strategy is adopted to evaluate the generalization ability of all the methods. The hyperparameters in the methods are optimized using grid search in the range as described in Table II. While five metrics are used to measure the performance of the methods, it is unlikely for a method to excel for all the metrics at the same time [1, 49-51]. Hence, only one metric is chosen as a reference to determine the corresponding optimal parameter values, although we can obtain different optimal parameter settings for a certain method on a specific dataset with different metrics. For example, Tables III and IV show the results under different optimal parameter settings that are determined with the metrics AP and HL respectively.

---

[1] https://github.com/KKimura360/MLC_toolbox/tree/master/dataset/matfile
[2] http://mulan.sourceforge.net/datasets-mlc.html

TABLE II
PARAMETER SETTINGS OF EXPERIMENTS

| Methods | Parameters ranges for grid search |
|---|---|
| ML-kNN | $K = \{1,3,5,7,9,11,13\}$, $s = \{0.01,0.03,0.05,0.07,0.09\}$. |
| MLSF | $\alpha = \{0.1,0.5,0.9\}$, $\varepsilon = \{0.1,1,10\}$, $K = \{1,5,10\}$, $\gamma = \{0.1,1,10\}$. |
| HNOML | $\alpha = \{0.01,0.1,1,10,100\}$, $\beta = \{0.01,0.1,1,10,100\}$, $\gamma = \{0.01,0.1,1,10,100\}$. |
| C2AE | $\alpha = \{0.1,0.3,0.5,0.7,0.9,1,3,5,7,9,10\}$, $N = 512$, $B = 500$. |
| BP-MLL | $e = \{10,20,30,40,50,60,70,80,90,100\}$, $h = 0.2*D$, $D$ is the number of features. |
| JBNN | $e = \{10,20,30,40,50,60,70,80,90,100\}$, $h = 0.2*D$, $D$ is the number of features. |
| BR | $C = 2.\wedge(-5:2:5)$, $M = \{2,3,4,5,6,7,8,9\}$. |
| CC | $C = 2.\wedge(-5:2:5)$, $M = \{2,3,4,5,6,7,8,9\}$. |
| **ML-TSK FS** | $K = \{1,2,3,4,5,6,7,8,9,10\}$, $h = \{0.1,1,10,100\}$, $\alpha = \{0.01,0.1,1,10,100\}$, $\beta = \{0.01,0.1,1,10,100\}$, $\gamma = \{0.1,1,10,100\}$. |

Due to the high computational complexity of BR, PCA is adopted to reduce the Mirflickr dataset to 50 dimensions. Similarly, PCA is employed in CC to reduce the Corel16k1 dataset to 150 dimensions.

*D. Analysis of Experimental Results*

*1) Multi-Label Classification Results*

The classification performance of the 9 methods on the 12 benchmark multi-label datasets, in terms of the mean and standard deviation (SD) of the five metrics, are shown in Tables III-V.

Multi-label method is usually evaluated from different aspects using different metrics, it is unusual and difficult for one method to outperform the others on every aspect evaluated with the corresponding metrics [1, 49-51]. In the study, we determine the optimal parameter setting using grid search based on one specific metric out of the five at a time. The results obtained using AP and HL respectively as the metric are discussed below.

The results obtained with AP adopted as the metric are shown in Table III. Through grid search, the parameters of the methods are optimized when AP reaches the maximum value. However, the other four metrics, i.e., HL, OE, RL and CV, are not necessarily optimized under such parameters settings. Therefore, as shown in Table III, the proposed ML-TSK FS does not demonstrate an overall advantage for the other methods, e.g., for the CAL500, Image and Scene datasets. Nevertheless, ML-TSK FS shows the best performance in terms of AP for 11 out of 12 datasets. It is only suboptimal for the Scene dataset.

Similarly, when HL is adopted as the metric to determine the optimal parameter settings for the methods using grid search, the settings thus obtained when HL is minimized do not necessarily guarantee that the metrics AP, OE, RL and CV are also optimized, as shown in the results in Table IV. Nevertheless, ML-TSK FS demonstrates outstanding performance in terms of HL for 10 out of the 12 datasets. For the Scene and Yeast datasets, the HL values of the proposed ML-TSK FS are the second best.

To fully demonstrate the performance of the methods based on different metrics, we have conducted further experiments to obtain the best values of the five metrics attained by the 9 methods on the 12 datasets separately. The results are shown in Table V, which shows that the proposed ML-TSK FS outperforms for each of the five metrics. This suggests the fuzzy inference relationship between features and labels are effectively modeled and the correlation between the labels is well utilized by the proposed ML-TSK FS.

Although CC and BR are two representative problem transformation methods, their classification performance, as shown in Tables III-V, is not outstanding. This implies that the transformation of multi-label problem into multiple single label problems in BR is not an effective solution and it ignores the correlation between the labels. On the other hand, although CC considers the correlation among labels, the classification results are dependent on the ordering of the labels.

*2) Label Correlation Analysis*

To evaluate whether the correlation between two labels is consistent with the correlation between their discriminative features, two experiments are conducted using the datasets Emotions and Scene. The results are shown with Fig. 3 and Fig. 4. The visualization of the correlation between any two optimized consequent parameter vectors $p_{g,i}$ $(1 \leq i \leq L)$ and $p_{g,j}$ $(1 \leq j \leq L)$ is given in Fig. 3, whereas the visualization of the correlation between any two labels $\tilde{y}_i$ and $\tilde{y}_j$ is given in Fig. 4. In these two figures, Pearson correlation coefficient is used to measure the correlation between any two columns in a matrix. The higher the value of the coefficient, the stronger the correlation. The effectiveness of label correlation learning based on fuzzy inference is analyzed from two aspects: (1) whether the correlation between two labels (expressed by $Y$ in Fig. 4) is consistent with the correlation between their discriminative features (expressed by optimized $P$ in Fig. 3); (2) whether the correlation expressed by $P$ is reasonable in the real situation. Comparing Fig. 3 with Fig. 4, the following conclusions can be drawn:

(1) By comparing Fig. 3 with Fig. 4, it is clear that the correlation between two labels is consistent with the correlation between their discriminative features.

(2) It can be seen from Fig. 3 that the results measured by $P$ are consistent with the real situation. For example, for the Emotions dataset, it is likely that "Amazed" is probably due to "Happy" or "Angry" rather than "Relaxing"; and that "Quiet" is probably due to "Sad", not "Happy". For the Scene dataset,

TABLE III
MEAN (SD) OF THE PEROFMRANCE METRICS UNDER THE OPTIMAL PARAMETERS SETTINGS OBTAINED WITH AP AS REFERENCE

| Method | Metrics | ML-kNN | HNOML | MLSF | CC | BR | C2AE | BP-MLL | JBNN | ML-TSK FS |
|---|---|---|---|---|---|---|---|---|---|---|
| Bibtex | AP | 0.35(0.01) | 0.58(0.01) | 0.37(0.02) | 0.58(0.01) | 0.60(0.01) | 0.09(0.03) | 0.54(0.01) | 0.02(0.00) | **0.61(0.00)** |
| | HL | 0.01(0.00) | 0.01(0.00) | 0.01(0.00) | 0.01(0.00) | 0.01(0.00) | 0.14(0.02) | 0.02(0.00) | 0.02(0.00) | **0.01(0.00)** |
| | OE | 0.59(0.01) | 0.37(0.01) | 0.57(0.01) | 0.37(0.02) | 0.36(0.01) | 0.94(0.04) | 0.42(0.02) | 0.98(0.00) | **0.35(0.01)** |
| | RL | 0.21(0.01) | 0.09(0.00) | 0.14(0.01) | 0.09(0.00) | 0.08(0.00) | 0.45(0.03) | 0.12(0.02) | 0.97(0.00) | **0.07(0.00)** |
| | CV | 0.34(0.01) | 0.17(0.01) | 0.35(0.02) | 0.18(0.00) | 0.16(0.00) | 0.60(0.03) | 0.17(0.01) | 0.62(0.01) | **0.14(0.01)** |
| Birds | AP | 0.22(0.02) | 0.34(0.03) | 0.26(0.03) | 0.34(0.01) | 0.33(0.03) | 0.30(0.04) | 0.34(0.02) | 0.29(0.06) | **0.34(0.03)** |
| | HL | 0.05(0.00) | 0.05(0.00) | 0.05(0.01) | 0.05(0.00) | 0.07(0.01) | 0.15(0.01) | 0.20(0.02) | 0.05(0.01) | **0.05(0.01)** |
| | OE | 0.85(0.03) | 0.67(0.05) | 0.79(0.03) | 0.67(0.02) | 0.67(0.02) | 0.96(0.01) | 0.72(0.04) | 0.94(0.03) | **0.67(0.04)** |
| | RL | 0.16(0.01) | 0.10(0.02) | **0.08(0.02)** | 0.10(0.02) | 0.11(0.02) | 0.24(0.03) | 0.10(0.01) | 0.39(0.06) | 0.10(0.02) |
| | CV | 0.19(0.01) | **0.12(0.02)** | 0.17(0.05) | 0.14(0.03) | 0.14(0.02) | 0.25(0.04) | 0.13(0.02) | 0.23(0.04) | 0.13(0.02) |
| CAL500 | AP | 0.50(0.01) | 0.43(0.18) | 0.49(0.01) | 0.46(0.01) | 0.50(0.01) | 0.33(0.02) | 0.46(0.02) | 0.45(0.01) | **0.52(0.01)** |
| | HL | 0.14(0.00) | 0.14(0.00) | 0.14(0.00) | 0.14(0.00) | 0.14(0.00) | 0.22(0.00) | 0.29(0.00) | 0.14(0.00) | **0.14(0.00)** |
| | OE | **0.12(0.03)** | 0.30(0.39) | 0.12(0.05) | 0.18(0.04) | 0.14(0.03) | 0.12(0.04) | 0.42(0.15) | 0.12(0.04) | 0.13(0.03) |
| | RL | 0.18(0.00) | **0.14(0.08)** | 0.18(0.01) | 0.23(0.01) | 0.19(0.01) | 0.36(0.01) | 0.19(0.00) | 0.28(0.01) | 0.18(0.00) |
| | CV | **0.75(0.01)** | 0.78(0.06) | 0.77(0.02) | 0.89(0.02) | 0.80(0.01) | 0.94(0.01) | 0.77(0.01) | 0.92(0.01) | 0.76(0.02) |
| Corel16k1 | AP | 0.28(0.00) | 0.34(0.01) | 0.28(0.01) | 0.30(0.00) | 0.34(0.00) | 0.21(0.02) | 0.26(0.01) | 0.08(0.00) | **0.35(0.01)** |
| | HL | 0.02(0.00) | **0.02(0.00)** | 0.02(0.00) | 0.02(0.00) | 0.02(0.00) | 0.21(0.02) | 0.15(0.00) | 0.02(0.00) | 0.02(0.00) |
| | OE | 0.74(0.01) | 0.65(0.01) | 0.73(0.01) | 0.71(0.01) | 0.65(0.00) | 0.80(0.03) | 0.82(0.02) | 0.86(0.02) | **0.63(0.01)** |
| | RL | 0.17(0.00) | 0.15(0.00) | **0.14(0.01)** | 0.16(0.00) | 0.16(0.00) | 0.31(0.02) | 0.15(0.00) | 0.75(0.02) | 0.16(0.00) |
| | CV | 0.33(0.00) | 0.31(0.00) | 0.39(0.02) | 0.33(0.01) | 0.31(0.01) | 0.53(0.03) | 0.30(0.00) | 0.72(0.01) | **0.30(0.00)** |
| Emotions | AP | 0.71(0.02) | 0.80(0.03) | 0.76(0.02) | 0.78(0.00) | 0.80(0.01) | 0.57(0.03) | 0.80(0.01) | 0.76(0.02) | **0.82(0.01)** |
| | HL | 0.26(0.01) | 0.21(0.02) | 0.24(0.02) | 0.22(0.02) | 0.21(0.02) | 0.41(0.03) | 0.22(0.01) | 0.20(0.00) | **0.19(0.01)** |
| | OE | 0.37(0.03) | 0.26(0.04) | 0.34(0.06) | 0.32(0.01) | 0.26(0.02) | 0.60(0.09) | 0.29(0.02) | 0.31(0.02) | **0.25(0.02)** |
| | RL | 0.27(0.02) | 0.17(0.02) | **0.11(0.01)** | 0.18(0.01) | 0.17(0.02) | 0.43(0.03) | 0.16(0.01) | 0.23(0.03) | 0.15(0.02) |
| | CV | 0.39(0.02) | 0.30(0.02) | 0.34(0.03) | 0.31(0.02) | 0.30(0.03) | 0.32(0.03) | 0.38(0.02) | 0.21(0.03) | **0.29(0.02)** |
| Flags | AP | 0.80(0.04) | 0.81(0.01) | 0.82(0.03) | 0.80(0.04) | 0.81(0.04) | 0.74(0.06) | 0.82(0.02) | 0.80(0.04) | **0.82(0.01)** |
| | HL | 0.33(0.04) | 0.28(0.02) | 0.28(0.04) | 0.28(0.04) | 0.27(0.03) | 0.42(0.03) | 0.33(0.02) | 0.30(0.03) | **0.26(0.03)** |
| | OE | 0.19(0.08) | 0.21(0.04) | 0.20(0.06) | 0.21(0.08) | **0.18(0.10)** | 0.32(0.13) | 0.20(0.03) | 0.23(0.07) | 0.19(0.01) |
| | RL | 0.24(0.04) | 0.22(0.01) | **0.13(0.02)** | 0.23(0.05) | 0.23(0.04) | 0.36(0.08) | 0.21(0.03) | 0.23(0.04) | 0.22(0.01) |
| | CV | 0.56(0.02) | **0.54(0.03)** | 0.55(0.04) | 0.56(0.02) | 0.56(0.02) | 0.54(0.03) | 0.49(0.04) | 0.53(0.03) | 0.55(0.01) |
| Image | AP | 0.74(0.02) | 0.78(0.02) | 0.72(0.02) | 0.78(0.03) | 0.79(0.03) | 0.47(0.02) | 0.79(0.02) | 0.63(0.04) | **0.79(0.03)** |
| | HL | 0.21(0.01) | 0.25(0.01) | 0.22(0.02) | 0.19(0.03) | **0.18(0.01)** | 0.46(0.04) | 0.21(0.01) | 0.21(0.00) | 0.19(0.01) |
| | OE | 0.40(0.04) | 0.34(0.04) | 0.43(0.04) | 0.35(0.05) | 0.34(0.06) | 0.81(0.03) | 0.34(0.03) | 0.53(0.06) | **0.33(0.05)** |
| | RL | 0.22(0.01) | 0.18(0.02) | **0.11(0.01)** | 0.18(0.03) | 0.17(0.02) | 0.52(0.04) | 0.17(0.02) | 0.37(0.05) | 0.17(0.03) |
| | CV | 0.23(0.01) | 0.20(0.02) | 0.25(0.01) | 0.20(0.03) | 0.19(0.02) | 0.24(0.04) | 0.21(0.02) | **0.18(0.04)** | 0.19(0.02) |
| Mirflickr | AP | 0.51(0.00) | 0.51(0.00) | 0.27(0.00) | 0.48(0.00) | 0.44(0.04) | 0.45(0.02) | 0.47(0.02) | 0.42(0.03) | **0.53(0.00)** |
| | HL | 0.15(0.00) | 0.15(0.00) | 0.15(0.00) | 0.16(0.00) | 0.16(0.01) | 0.30(0.03) | 0.31(0.00) | 0.15(0.00) | **0.15(0.00)** |
| | OE | 0.53(0.01) | 0.50(0.01) | **0.02(0.00)** | 0.57(0.00) | 0.58(0.05) | 0.66(0.04) | 0.66(0.03) | 0.51(0.01) | 0.49(0.01) |
| | RL | 0.21(0.00) | 0.21(0.00) | 0.70(0.00) | 0.24(0.00) | 0.32(0.04) | 0.25(0.01) | 0.21(0.01) | 0.53(0.05) | **0.20(0.00)** |
| | CV | 0.44(0.00) | 0.44(0.00) | 0.64(0.00) | 0.52(0.01) | 0.62(0.04) | 0.46(0.02) | 0.40(0.00) | 0.60(0.01) | **0.43(0.00)** |
| Rcv1s1 | AP | 0.49(0.01) | 0.61(0.01) | 0.52(0.02) | 0.57(0.01) | 0.60(0.01) | 0.21(0.02) | 0.53(0.05) | 0.05(0.01) | **0.61(0.00)** |
| | HL | 0.03(0.00) | 0.03(0.00) | 0.03(0.00) | 0.03(0.00) | 0.03(0.00) | 0.17(0.02) | 0.04(0.00) | 0.03(0.00) | **0.03(0.00)** |
| | OE | 0.54(0.01) | 0.42(0.01) | 0.51(0.01) | 0.47(0.01) | 0.42(0.01) | 0.78(0.06) | 0.60(0.14) | 0.96(0.01) | **0.42(0.01)** |
| | RL | 0.09(0.00) | **0.04(0.00)** | 0.08(0.02) | 0.07(0.00) | 0.06(0.00) | 0.33(0.02) | 0.08(0.01) | 0.90(0.02) | 0.05(0.00) |
| | CV | 0.20(0.00) | **0.11(0.00)** | 0.24(0.04) | 0.19(0.01) | 0.14(0.01) | 0.51(0.03) | 0.15(0.01) | 0.67(0.01) | 0.12(0.01) |
| Rcv1s2 | AP | 0.50(0.01) | 0.63(0.00) | 0.52(0.02) | 0.58(0.01) | 0.61(0.01) | 0.18(0.04) | 0.58(0.01) | 0.05(0.01) | **0.64(0.01)** |
| | HL | 0.02(0.00) | 0.02(0.00) | 0.02(0.00) | 0.02(0.00) | 0.02(0.00) | 0.17(0.03) | 0.03(0.00) | 0.03(0.00) | **0.02(0.00)** |
| | OE | 0.56(0.02) | 0.42(0.02) | 0.54(0.02) | 0.46(0.01) | 0.44(0.01) | 0.83(0.10) | 0.47(0.02) | 0.96(0.02) | **0.40(0.02)** |
| | RL | 0.09(0.00) | 0.05(0.00) | 0.06(0.00) | 0.07(0.00) | 0.06(0.00) | 0.36(0.02) | 0.09(0.01) | 0.88(0.02) | **0.05(0.00)** |
| | CV | 0.19(0.01) | 0.13(0.01) | 0.19(0.01) | 0.18(0.01) | 0.14(0.01) | 0.55(0.02) | 0.15(0.01) | 0.62(0.01) | **0.12(0.01)** |
| Scene | AP | **0.87(0.01)** | 0.85(0.01) | 0.86(0.02) | 0.84(0.01) | 0.86(0.01) | 0.42(0.01) | 0.87(0.01) | 0.59(0.03) | 0.86(0.01) |
| | HL | 0.09(0.00) | 0.13(0.00) | 0.09(0.00) | 0.10(0.01) | 0.11(0.01) | 0.41(0.03) | 0.11(0.00) | 0.14(0.00) | 0.11(0.01) |
| | OE | **0.23(0.01)** | 0.25(0.03) | 0.23(0.04) | 0.27(0.02) | 0.25(0.03) | 0.81(0.02) | 0.22(0.01) | 0.54(0.02) | 0.24(0.02) |
| | RL | 0.08(0.01) | 0.08(0.01) | **0.04(0.00)** | 0.09(0.01) | 0.09(0.02) | 0.49(0.01) | 0.07(0.00) | 0.41(0.05) | 0.08(0.01) |
| | CV | 0.08(0.01) | 0.08(0.01) | 0.08(0.01) | 0.09(0.01) | 0.09(0.02) | 0.26(0.01) | 0.09(0.00) | 0.15(0.03) | **0.08(0.01)** |
| Yeast | AP | 0.76(0.02) | 0.61(0.00) | 0.75(0.02) | 0.72(0.01) | 0.76(0.01) | 0.57(0.02) | 0.74(0.01) | 0.71(0.03) | **0.76(0.01)** |
| | HL | **0.19(0.01)** | 0.30(0.00) | 0.20(0.00) | 0.21(0.01) | 0.20(0.00) | 0.34(0.04) | 0.23(0.01) | 0.21(0.01) | 0.20(0.01) |
| | OE | 0.23(0.02) | 0.36(0.01) | 0.25(0.03) | 0.26(0.03) | 0.23(0.02) | 0.62(0.03) | 0.25(0.01) | 0.25(0.01) | **0.21(0.01)** |
| | RL | 0.17(0.01) | 0.34(0.00) | **0.13(0.01)** | 0.21(0.01) | 0.17(0.00) | 0.30(0.02) | 0.18(0.01) | 0.23(0.03) | 0.17(0.01) |
| | CV | 0.45(0.02) | 0.62(0.01) | 0.48(0.01) | 0.52(0.01) | **0.45(0.01)** | 0.48(0.02) | 0.47(0.01) | 0.48(0.05) | 0.46(0.01) |

TABLE IV
MEAN (SD) OF THE PERFORMANCE METRICS UNDER THE OPTIMAL PARAMETERS SETTINGS OBTAINED WITH HL AS REFERENCE

| Method | Metrics | ML-kNN | HNOML | MLSF | CC | BR | C2AE | BP-MLL | JBNN | ML-TSK FS |
|---|---|---|---|---|---|---|---|---|---|---|
| Bibtex | HL | 0.01(0.00) | 0.01(0.00) | 0.01(0.00) | 0.01(0.00) | 0.01(0.00) | 0.14(0.02) | 0.02(0.00) | 0.02(0.00) | **0.01(0.00)** |
| | AP | 0.33(0.01) | 0.57(0.01) | 0.36(0.02) | 0.58(0.01) | 0.60(0.01) | 0.08(0.03) | 0.54(0.01) | 0.02(0.00) | **0.60(0.01)** |
| | OE | 0.61(0.01) | 0.37(0.01) | 0.57(0.02) | 0.37(0.02) | 0.36(0.01) | 0.96(0.03) | 0.42(0.02) | 0.98(0.00) | **0.35(0.01)** |
| | RL | 0.23(0.01) | 0.10(0.00) | 0.14(0.01) | 0.09(0.00) | 0.09(0.00) | 0.43(0.02) | 0.12(0.02) | 0.99(0.00) | **0.08(0.00)** |
| | CV | 0.38(0.01) | 0.18(0.01) | 0.37(0.01) | 0.18(0.00) | 0.17(0.00) | 0.58(0.01) | 0.17(0.01) | 0.62(0.01) | **0.16(0.01)** |
| Birds | HL | 0.05(0.00) | 0.05(0.00) | 0.05(0.01) | 0.05(0.00) | 0.06(0.00) | 0.15(0.01) | 0.18(0.02) | 0.05(0.01) | **0.05(0.00)** |
| | AP | 0.22(0.02) | **0.34(0.03)** | 0.25(0.03) | 0.33(0.03) | 0.32(0.02) | 0.27(0.04) | 0.29(0.03) | 0.28(0.04) | 0.31(0.05) |
| | OE | 0.84(0.04) | **0.66(0.02)** | 0.79(0.03) | 0.66(0.02) | 0.69(0.02) | 0.96(0.01) | 0.73(0.02) | 0.96(0.03) | 0.73(0.06) |
| | RL | 0.16(0.01) | 0.09(0.02) | **0.08(0.02)** | 0.11(0.02) | 0.12(0.01) | 0.31(0.03) | 0.10(0.01) | 0.37(0.04) | 0.11(0.03) |
| | CV | 0.19(0.01) | **0.12(0.03)** | 0.18(0.04) | 0.15(0.03) | 0.15(0.02) | 0.31(0.04) | 0.13(0.02) | 0.27(0.03) | 0.14(0.03) |
| CAL500 | HL | 0.14(0.01) | 0.14(0.01) | 0.14(0.00) | 0.14(0.00) | 0.14(0.00) | 0.19(0.01) | 0.29(0.01) | 0.14(0.00) | **0.14(0.00)** |
| | AP | 0.50(0.01) | 0.43(0.18) | 0.12(0.00) | 0.46(0.00) | 0.50(0.01) | 0.10(0.12) | 0.45(0.02) | 0.45(0.01) | **0.52(0.01)** |
| | OE | **0.12(0.03)** | 0.30(0.39) | 0.12(0.05) | 0.16(0.03) | 0.14(0.03) | 0.35(0.42) | 0.50(0.11) | 0.34(0.02) | 0.13(0.02) |
| | RL | 0.18(0.00) | **0.14(0.08)** | 0.95(0.00) | 0.24(0.01) | 0.19(0.01) | 0.16(0.19) | 0.19(0.01) | 0.28(0.02) | 0.18(0.01) |
| | CV | 0.75(0.01) | 0.78(0.06) | 0.98(0.00) | 0.90(0.01) | 0.80(0.01) | 0.79(0.08) | 0.77(0.01) | 0.91(0.01) | **0.75(0.02)** |
| Corel16k1 | HL | 0.02(0.00) | 0.02(0.00) | 0.02(0.00) | 0.02(0.00) | 0.02(0.00) | 0.21(0.02) | 0.12(0.00) | 0.02(0.00) | **0.02(0.00)** |
| | AP | 0.28(0.00) | 0.34(0.01) | 0.27(0.02) | 0.30(0.00) | 0.34(0.00) | 0.19(0.01) | 0.21(0.00) | 0.08(0.00) | **0.34(0.01)** |
| | OE | 0.74(0.01) | 0.65(0.01) | 0.74(0.01) | 0.71(0.01) | 0.65(0.00) | 0.81(0.01) | 0.80(0.01) | 0.86(0.02) | **0.63(0.01)** |
| | RL | 0.17(0.00) | **0.15(0.00)** | 0.16(0.03) | 0.16(0.00) | 0.16(0.00) | 0.31(0.02) | 0.16(0.00) | 0.75(0.02) | 0.16(0.01) |
| | CV | 0.34(0.00) | 0.31(0.00) | 0.43(0.08) | 0.33(0.01) | 0.31(0.01) | 0.54(0.03) | **0.30(0.00)** | 0.72(0.01) | 0.32(0.01) |
| Emotions | HL | 0.26(0.01) | 0.21(0.01) | 0.24(0.02) | 0.21(0.02) | 0.20(0.01) | 0.41(0.03) | 0.22(0.02) | 0.20(0.00) | **0.19(0.01)** |
| | AP | 0.71(0.01) | 0.80(0.03) | 0.73(0.04) | 0.78(0.02) | 0.80(0.02) | 0.56(0.03) | 0.79(0.01) | 0.76(0.02) | **0.81(0.01)** |
| | OE | 0.39(0.01) | 0.26(0.05) | 0.38(0.03) | 0.33(0.02) | 0.29(0.03) | 0.60(0.09) | 0.31(0.00) | 0.31(0.02) | **0.25(0.03)** |
| | RL | 0.27(0.02) | 0.17(0.02) | **0.12(0.03)** | 0.19(0.03) | 0.17(0.02) | 0.45(0.03) | 0.16(0.01) | 0.23(0.03) | 0.15(0.02) |
| | CV | 0.39(0.02) | 0.31(0.02) | 0.37(0.07) | 0.33(0.04) | 0.30(0.03) | 0.35(0.03) | 0.37(0.01) | 0.21(0.03) | **0.29(0.03)** |
| Flags | HL | 0.33(0.03) | 0.27(0.01) | 0.26(0.05) | 0.27(0.03) | 0.27(0.03) | 0.42(0.03) | 0.30(0.04) | 0.30(0.01) | **0.26(0.03)** |
| | AP | 0.78(0.03) | 0.80(0.02) | 0.80(0.03) | 0.80(0.03) | 0.81(0.04) | 0.70(0.02) | 0.80(0.03) | 0.77(0.02) | **0.82(0.01)** |
| | OE | 0.29(0.09) | 0.20(0.04) | 0.23(0.05) | 0.22(0.09) | 0.20(0.10) | 0.28(0.11) | 0.25(0.10) | 0.34(0.09) | **0.19(0.01)** |
| | RL | 0.25(0.03) | 0.23(0.02) | **0.13(0.02)** | 0.24(0.04) | 0.23(0.04) | 0.40(0.02) | 0.22(0.02) | 0.26(0.02) | 0.22(0.01) |
| | CV | 0.57(0.03) | 0.55(0.02) | 0.55(0.03) | 0.56(0.02) | 0.56(0.02) | 0.56(0.03) | 0.50(0.03) | 0.57(0.03) | **0.55(0.01)** |
| Image | HL | 0.20(0.01) | 0.23(0.02) | 0.21(0.01) | 0.19(0.03) | 0.18(0.01) | 0.46(0.04) | 0.21(0.01) | 0.21(0.00) | **0.18(0.01)** |
| | AP | 0.74(0.02) | 0.74(0.01) | 0.51(0.01) | 0.78(0.03) | 0.79(0.03) | 0.46(0.02) | **0.79(0.02)** | 0.63(0.04) | 0.78(0.01) |
| | OE | 0.41(0.03) | 0.40(0.03) | 0.43(0.04) | 0.35(0.05) | **0.32(0.05)** | 0.80(0.03) | 0.33(0.03) | 0.53(0.06) | 0.35(0.02) |
| | RL | 0.22(0.02) | 0.22(0.01) | 0.42(0.00) | 0.18(0.03) | **0.17(0.02)** | 0.52(0.02) | 0.17(0.02) | 0.37(0.05) | 0.19(0.00) |
| | CV | 0.23(0.02) | 0.23(0.02) | 0.42(0.02) | 0.20(0.03) | 0.20(0.01) | 0.26(0.02) | 0.23(0.02) | 0.21(0.04) | **0.19(0.02)** |
| Mirflickr | HL | 0.15(0.00) | 0.15(0.00) | 0.15(0.00) | 0.16(0.00) | 0.16(0.01) | 0.30(0.03) | 0.31(0.00) | 0.15(0.00) | **0.15(0.00)** |
| | AP | 0.51(0.00) | 0.51(0.00) | 0.27(0.00) | 0.47(0.01) | 0.44(0.04) | 0.44(0.02) | 0.46(0.04) | 0.40(0.04) | **0.53(0.00)** |
| | OE | 0.53(0.01) | 0.50(0.01) | **0.02(0.00)** | 0.58(0.01) | 0.58(0.05) | 0.68(0.11) | 0.65(0.01) | 0.51(0.01) | 0.49(0.01) |
| | RL | 0.21(0.00) | 0.21(0.00) | 0.70(0.00) | 0.25(0.00) | 0.32(0.04) | 0.25(0.01) | 0.21(0.01) | 0.60(0.08) | **0.20(0.00)** |
| | CV | 0.44(0.00) | 0.44(0.00) | 0.64(0.00) | 0.53(0.01) | 0.62(0.04) | 0.46(0.02) | 0.39(0.00) | 0.61(0.02) | **0.43(0.00)** |
| Rcv1s1 | HL | 0.03(0.00) | 0.03(0.00) | 0.03(0.00) | 0.03(0.00) | 0.03(0.00) | 0.17(0.02) | 0.04(0.00) | 0.03(0.00) | **0.03(0.00)** |
| | AP | 0.48(0.01) | 0.60(0.01) | 0.50(0.02) | 0.57(0.01) | 0.60(0.01) | 0.21(0.02) | 0.51(0.06) | 0.05(0.01) | **0.61(0.01)** |
| | OE | 0.54(0.01) | 0.43(0.01) | 0.52(0.02) | 0.47(0.01) | 0.43(0.01) | 0.78(0.06) | 0.58(0.13) | 0.96(0.01) | **0.42(0.02)** |
| | RL | 0.11(0.00) | 0.06(0.00) | 0.09(0.02) | 0.07(0.00) | 0.06(0.00) | 0.33(0.02) | 0.09(0.01) | 0.90(0.02) | **0.05(0.00)** |
| | CV | 0.22(0.01) | 0.14(0.01) | 0.27(0.04) | 0.19(0.01) | 0.14(0.01) | 0.51(0.03) | 0.16(0.01) | 0.67(0.01) | **0.13(0.00)** |
| Rcv1s2 | HL | 0.02(0.00) | 0.02(0.00) | 0.02(0.00) | 0.02(0.00) | 0.02(0.00) | 0.17(0.03) | 0.03(0.00) | 0.03(0.00) | **0.02(0.00)** |
| | AP | 0.49(0.01) | 0.62(0.00) | 0.51(0.02) | 0.58(0.01) | 0.61(0.01) | 0.18(0.04) | 0.58(0.01) | 0.05(0.01) | **0.62(0.00)** |
| | OE | 0.57(0.01) | 0.42(0.02) | 0.54(0.02) | 0.47(0.01) | 0.44(0.02) | 0.83(0.10) | 0.47(0.02) | 0.96(0.02) | **0.41(0.01)** |
| | RL | 0.10(0.00) | 0.07(0.00) | 0.09(0.02) | 0.08(0.00) | 0.06(0.00) | 0.36(0.02) | 0.09(0.01) | 0.88(0.02) | **0.06(0.00)** |
| | CV | 0.21(0.01) | 0.15(0.01) | 0.25(0.02) | 0.18(0.01) | 0.14(0.01) | 0.55(0.02) | 0.15(0.01) | 0.62(0.01) | **0.13(0.01)** |
| Scene | HL | **0.09(0.00)** | 0.12(0.00) | 0.09(0.01) | 0.10(0.01) | 0.10(0.01) | 0.41(0.03) | 0.11(0.00) | 0.14(0.00) | 0.10(0.01) |
| | AP | **0.87(0.01)** | 0.83(0.01) | 0.84(0.05) | 0.84(0.01) | 0.86(0.01) | 0.42(0.01) | 0.87(0.01) | 0.59(0.03) | 0.85(0.01) |
| | OE | **0.23(0.01)** | 0.27(0.02) | 0.25(0.05) | 0.27(0.02) | 0.23(0.02) | 0.81(0.02) | 0.22(0.01) | 0.54(0.02) | 0.24(0.02) |
| | RL | 0.08(0.01) | 0.10(0.01) | **0.05(0.03)** | 0.09(0.01) | 0.08(0.01) | 0.49(0.01) | 0.07(0.00) | 0.41(0.05) | 0.09(0.01) |
| | CV | **0.08(0.01)** | 0.10(0.01) | 0.11(0.07) | 0.09(0.01) | 0.08(0.01) | 0.26(0.01) | 0.09(0.00) | 0.15(0.03) | 0.09(0.01) |
| Yeast | HL | 0.19(0.01) | 0.30(0.00) | 0.19(0.00) | 0.21(0.01) | **0.19(0.00)** | 0.34(0.04) | 0.22(0.01) | 0.21(0.01) | 0.20(0.01) |
| | AP | 0.76(0.02) | 0.60(0.01) | 0.41(0.01) | 0.72(0.01) | 0.76(0.01) | 0.57(0.02) | 0.74(0.01) | 0.71(0.03) | **0.76(0.01)** |
| | OE | 0.24(0.02) | 0.37(0.02) | 0.26(0.03) | 0.26(0.03) | 0.23(0.02) | 0.62(0.03) | 0.25(0.01) | 0.25(0.01) | **0.22(0.03)** |
| | RL | 0.17(0.01) | 0.35(0.01) | 0.71(0.00) | 0.21(0.01) | **0.16(0.00)** | 0.30(0.02) | 0.18(0.01) | 0.25(0.04) | 0.17(0.01) |
| | CV | 0.45(0.02) | 0.62(0.01) | 0.77(0.01) | 0.52(0.01) | **0.44(0.01)** | 0.48(0.02) | 0.49(0.02) | 0.51(0.09) | 0.46(0.02) |

TABLE V
MEAN (SD) OF THE PEROFMRANCE METRICS OF THE MULTI-LABEL LEARNING METHODS

| Method | Metrics | ML-kNN | HNOML | MLSF | CC | BR | C2AE | BP-MLL | JBNN | ML-TSK FS |
|---|---|---|---|---|---|---|---|---|---|---|
| Bibtex | AP | 0.35(0.01) | 0.58(0.01) | 0.37(0.02) | 0.58(0.01) | 0.60(0.01) | 0.09(0.03) | 0.54(0.01) | 0.02(0.00) | **0.61(0.00)** |
| | HL | 0.01(0.00) | 0.01(0.00) | 0.01(0.00) | 0.01(0.00) | 0.01(0.00) | 0.14(0.02) | 0.02(0.00) | 0.02(0.00) | **0.01(0.00)** |
| | OE | 0.59(0.01) | 0.37(0.01) | 0.57(0.02) | 0.37(0.02) | 0.36(0.01) | 0.94(0.05) | 0.42(0.02) | 0.98(0.00) | **0.00(0.00)** |
| | RL | 0.21(0.01) | 0.07(0.00) | 0.14(0.01) | 0.09(0.00) | 0.08(0.00) | 0.42(0.01) | 0.08(0.01) | 0.97(0.00) | **0.06(0.01)** |
| | CV | 0.34(0.01) | 0.13(0.00) | 0.35(0.02) | 0.18(0.00) | 0.16(0.00) | 0.56(0.01) | 0.14(0.02) | 0.61(0.00) | **0.12(0.01)** |
| Birds | AP | 0.22(0.02) | 0.34(0.03) | 0.26(0.03) | 0.34(0.01) | 0.33(0.03) | 0.30(0.04) | 0.34(0.02) | 0.29(0.06) | **0.34(0.03)** |
| | HL | 0.05(0.00) | 0.05(0.00) | 0.05(0.01) | 0.05(0.00) | 0.06(0.00) | 0.15(0.01) | 0.18(0.02) | 0.05(0.01) | **0.05(0.00)** |
| | OE | 0.84(0.04) | 0.66(0.04) | 0.49(0.04) | 0.66(0.02) | 0.67(0.02) | 0.94(0.01) | 0.72(0.03) | 0.94(0.03) | **0.46(0.04)** |
| | RL | 0.16(0.01) | 0.09(0.02) | **0.08(0.02)** | 0.10(0.01) | 0.10(0.02) | 0.21(0.02) | 0.10(0.01) | 0.37(0.03) | 0.09(0.02) |
| | CV | 0.19(0.01) | 0.12(0.00) | 0.17(0.05) | 0.13(0.00) | 0.13(0.01) | 0.22(0.03) | 0.13(0.02) | 0.23(0.04) | **0.11(0.03)** |
| CAL500 | AP | 0.50(0.01) | 0.43(0.18) | 0.49(0.01) | 0.46(0.01) | 0.50(0.01) | 0.33(0.02) | 0.46(0.02) | 0.45(0.02) | **0.52(0.01)** |
| | HL | 0.14(0.00) | 0.14(0.00) | 0.14(0.00) | 0.14(0.00) | 0.14(0.00) | 0.19(0.01) | 0.29(0.00) | 0.14(0.00) | **0.14(0.00)** |
| | OE | 0.12(0.03) | 0.30(0.04) | 0.12(0.05) | 0.11(0.06) | 0.13(0.03) | 0.12(0.04) | 0.42(0.15) | 0.12(0.04) | **0.09(0.02)** |
| | RL | 0.18(0.00) | **0.14(0.08)** | 0.18(0.01) | 0.23(0.01) | 0.19(0.00) | 0.16(0.19) | 0.19(0.00) | 0.28(0.02) | 0.17(0.01) |
| | CV | 0.75(0.01) | 0.77(0.06) | 0.76(0.03) | 0.89(0.02) | 0.79(0.01) | 0.79(0.08) | 0.76(0.02) | 0.91(0.01) | **0.73(0.01)** |
| Corel16k1 | AP | 0.28(0.00) | 0.34(0.01) | 0.28(0.01) | 0.30(0.00) | 0.34(0.00) | 0.21(0.02) | 0.26(0.01) | 0.08(0.00) | **0.35(0.01)** |
| | HL | 0.02(0.00) | 0.02(0.00) | 0.02(0.00) | 0.02(0.00) | 0.02(0.00) | 0.21(0.02) | 0.12(0.00) | 0.02(0.00) | **0.02(0.00)** |
| | OE | 0.74(0.01) | 0.65(0.01) | 0.73(0.01) | 0.71(0.01) | 0.65(0.00) | 0.80(0.03) | 0.80(0.01) | 0.86(0.02) | **0.59(0.01)** |
| | RL | 0.17(0.00) | 0.15(0.00) | 0.14(0.01) | 0.16(0.00) | 0.16(0.00) | 0.30(0.02) | 0.15(0.00) | 0.75(0.02) | **0.14(0.00)** |
| | CV | 0.33(0.00) | 0.31(0.00) | 0.39(0.02) | 0.33(0.01) | 0.31(0.01) | 0.53(0.03) | 0.30(0.01) | 0.71(0.00) | **0.29(0.00)** |
| Emotions | AP | 0.71(0.02) | 0.80(0.03) | 0.76(0.02) | 0.78(0.00) | 0.80(0.01) | 0.57(0.03) | 0.80(0.01) | 0.76(0.02) | **0.82(0.01)** |
| | HL | 0.26(0.01) | 0.21(0.01) | 0.24(0.02) | 0.21(0.02) | 0.20(0.01) | 0.41(0.03) | 0.22(0.02) | 0.20(0.00) | **0.19(0.01)** |
| | OE | 0.37(0.03) | 0.26(0.05) | 0.34(0.06) | 0.31(0.03) | 0.26(0.02) | 0.60(0.09) | 0.29(0.02) | 0.31(0.01) | **0.20(0.05)** |
| | RL | 0.26(0.02) | 0.16(0.02) | **0.11(0.02)** | 0.18(0.01) | 0.17(0.02) | 0.43(0.03) | 0.16(0.01) | 0.23(0.03) | 0.15(0.03) |
| | CV | 0.38(0.02) | 0.30(0.02) | 0.33(0.03) | 0.31(0.03) | 0.30(0.03) | 0.32(0.03) | 0.37(0.01) | 0.20(0.03) | **0.28(0.03)** |
| Flags | AP | 0.80(0.04) | 0.81(0.01) | 0.82(0.03) | 0.80(0.04) | 0.81(0.04) | 0.74(0.06) | 0.82(0.02) | 0.80(0.04) | **0.82(0.01)** |
| | HL | 0.33(0.03) | 0.27(0.01) | 0.26(0.05) | 0.27(0.03) | 0.27(0.03) | 0.42(0.03) | 0.30(0.04) | 0.30(0.01) | **0.26(0.03)** |
| | OE | 0.19(0.08) | 0.20(0.04) | 0.20(0.08) | 0.19(0.08) | 0.18(0.10) | 0.21(0.07) | 0.20(0.05) | 0.20(0.05) | **0.17(0.02)** |
| | RL | 0.24(0.04) | 0.22(0.01) | **0.12(0.02)** | 0.23(0.05) | 0.22(0.04) | 0.36(0.08) | 0.21(0.03) | 0.23(0.04) | 0.21(0.02) |
| | CV | 0.56(0.02) | 0.54(0.03) | 0.54(0.04) | 0.56(0.03) | 0.55(0.02) | 0.54(0.02) | **0.48(0.04)** | 0.53(0.03) | 0.52(0.01) |
| Image | AP | 0.74(0.02) | 0.78(0.02) | 0.72(0.02) | 0.78(0.03) | 0.79(0.03) | 0.47(0.02) | 0.79(0.02) | 0.63(0.04) | **0.79(0.03)** |
| | HL | 0.20(0.01) | 0.23(0.02) | 0.21(0.01) | 0.19(0.03) | 0.18(0.01) | 0.46(0.04) | 0.21(0.01) | 0.21(0.00) | **0.18(0.01)** |
| | OE | 0.40(0.04) | 0.34(0.04) | 0.43(0.04) | 0.35(0.06) | 0.32(0.05) | 0.80(0.03) | 0.33(0.03) | 0.53(0.06) | **0.01(0.01)** |
| | RL | 0.22(0.02) | 0.18(0.02) | **0.10(0.01)** | 0.18(0.03) | 0.17(0.02) | 0.52(0.03) | 0.17(0.02) | 0.37(0.05) | 0.12(0.02) |
| | CV | 0.23(0.02) | 0.20(0.02) | 0.24(0.01) | 0.20(0.03) | 0.19(0.02) | 0.24(0.03) | 0.21(0.02) | **0.17(0.04)** | 0.18(0.02) |
| Mirflickr | AP | 0.51(0.00) | 0.51(0.00) | 0.27(0.00) | 0.48(0.00) | 0.44(0.04) | 0.45(0.02) | 0.47(0.02) | 0.42(0.03) | **0.53(0.00)** |
| | HL | 0.15(0.00) | 0.15(0.00) | 0.15(0.00) | 0.16(0.00) | 0.16(0.01) | 0.30(0.03) | 0.31(0.00) | 0.15(0.00) | **0.15(0.00)** |
| | OE | 0.53(0.01) | 0.50(0.01) | **0.02(0.00)** | 0.57(0.00) | 0.58(0.05) | 0.66(0.04) | 0.64(0.02) | 0.50(0.01) | 0.09(0.00) |
| | RL | 0.21(0.00) | 0.21(0.00) | 0.26(0.00) | 0.24(0.00) | 0.32(0.04) | 0.25(0.02) | 0.21(0.01) | 0.53(0.05) | **0.19(0.00)** |
| | CV | 0.44(0.00) | 0.44(0.00) | 0.45(0.00) | 0.52(0.01) | 0.62(0.04) | 0.46(0.02) | **0.39(0.00)** | 0.60(0.01) | 0.42(0.00) |
| Rcv1s1 | AP | 0.49(0.01) | 0.61(0.01) | 0.52(0.02) | 0.57(0.01) | 0.60(0.01) | 0.21(0.02) | 0.53(0.05) | 0.05(0.01) | **0.61(0.00)** |
| | HL | 0.03(0.00) | 0.03(0.00) | 0.03(0.00) | 0.03(0.00) | 0.03(0.00) | 0.17(0.02) | 0.04(0.00) | 0.03(0.00) | **0.03(0.00)** |
| | OE | 0.54(0.01) | 0.42(0.01) | 0.51(0.01) | 0.47(0.01) | 0.42(0.01) | 0.78(0.06) | 0.58(0.13) | 0.96(0.01) | **0.01(0.00)** |
| | RL | 0.09(0.00) | **0.04(0.00)** | 0.08(0.01) | 0.07(0.00) | 0.06(0.00) | 0.31(0.01) | 0.07(0.01) | 0.90(0.02) | 0.05(0.00) |
| | CV | 0.20(0.00) | 0.11(0.00) | 0.24(0.04) | 0.17(0.01) | 0.14(0.01) | 0.50(0.02) | 0.15(0.01) | 0.65(0.01) | **0.11(0.00)** |
| Rcv1s2 | AP | 0.50(0.01) | 0.63(0.00) | 0.52(0.02) | 0.58(0.01) | 0.61(0.01) | 0.18(0.04) | 0.58(0.01) | 0.05(0.01) | **0.64(0.01)** |
| | HL | 0.02(0.00) | 0.02(0.00) | 0.02(0.00) | 0.02(0.00) | 0.02(0.00) | 0.17(0.03) | 0.03(0.00) | 0.03(0.00) | **0.02(0.00)** |
| | OE | 0.56(0.01) | 0.41(0.02) | 0.53(0.01) | 0.46(0.01) | 0.44(0.01) | 0.81(0.06) | 0.47(0.02) | 0.96(0.02) | **0.07(0.11)** |
| | RL | 0.09(0.00) | **0.04(0.00)** | 0.06(0.00) | 0.07(0.00) | 0.06(0.00) | 0.35(0.03) | 0.07(0.01) | 0.88(0.02) | 0.05(0.00) |
| | CV | 0.19(0.01) | 0.11(0.00) | 0.19(0.01) | 0.16(0.01) | 0.14(0.01) | 0.53(0.04) | 0.14(0.01) | 0.62(0.01) | **0.10(0.01)** |
| Scene | AP | 0.87(0.01) | 0.85(0.01) | 0.86(0.02) | 0.84(0.01) | 0.86(0.01) | 0.42(0.01) | **0.87(0.01)** | 0.59(0.03) | 0.86(0.01) |
| | HL | **0.09(0.00)** | 0.12(0.00) | 0.09(0.00) | 0.10(0.01) | 0.10(0.01) | 0.41(0.03) | 0.11(0.00) | 0.14(0.00) | 0.10(0.01) |
| | OE | 0.23(0.01) | 0.25(0.03) | 0.23(0.04) | 0.27(0.02) | 0.23(0.02) | 0.81(0.02) | 0.22(0.01) | 0.54(0.02) | **0.00(0.00)** |
| | RL | 0.08(0.01) | 0.08(0.01) | **0.04(0.01)** | 0.09(0.01) | 0.08(0.01) | 0.49(0.01) | 0.07(0.00) | 0.41(0.04) | 0.07(0.02) |
| | CV | 0.08(0.01) | 0.08(0.01) | 0.08(0.01) | 0.09(0.01) | 0.08(0.01) | 0.26(0.01) | 0.09(0.00) | 0.14(0.03) | **0.08(0.01)** |
| Yeast | AP | 0.76(0.02) | 0.61(0.00) | 0.75(0.02) | 0.72(0.01) | 0.76(0.01) | 0.57(0.02) | 0.74(0.01) | 0.71(0.03) | **0.76(0.01)** |
| | HL | 0.19(0.01) | 0.30(0.00) | 0.19(0.00) | 0.21(0.01) | **0.19(0.00)** | 0.34(0.04) | 0.22(0.01) | 0.21(0.01) | 0.20(0.01) |
| | OE | 0.23(0.01) | 0.36(0.01) | 0.25(0.02) | 0.26(0.03) | 0.22(0.01) | 0.61(0.04) | 0.25(0.01) | 0.25(0.01) | **0.21(0.01)** |
| | RL | 0.17(0.01) | 0.34(0.00) | **0.13(0.01)** | 0.21(0.00) | 0.16(0.00) | 0.30(0.02) | 0.18(0.01) | 0.23(0.03) | 0.15(0.01) |
| | CV | 0.45(0.02) | 0.62(0.01) | 0.48(0.03) | 0.51(0.02) | 0.44(0.01) | 0.47(0.03) | 0.47(0.01) | 0.48(0.05) | **0.42(0.01)** |

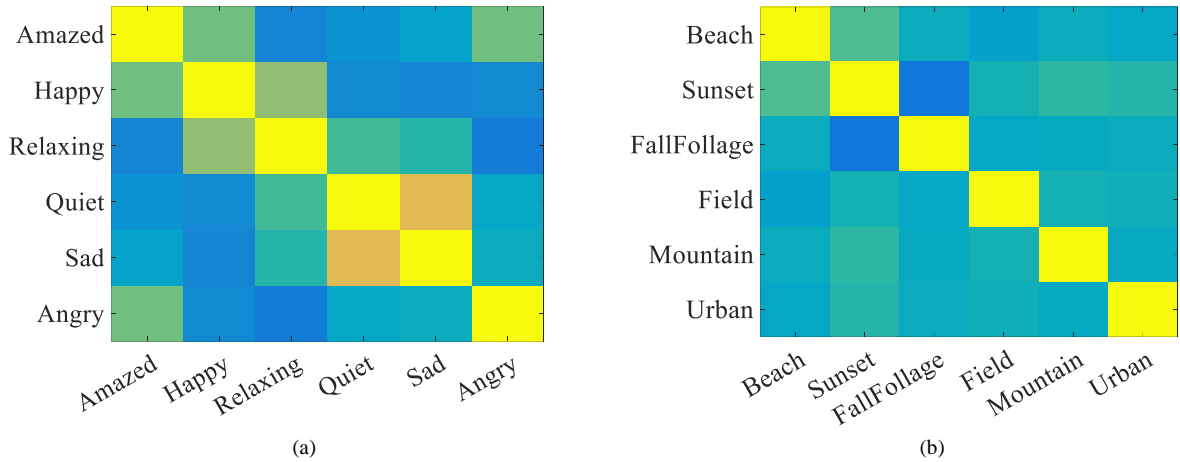

Fig. 3 Visualization of the correlation between any two sets of discriminative features (expressed by the consequent parameter matrix **P**) on the (a) Emotions and (b) Scene datasets.

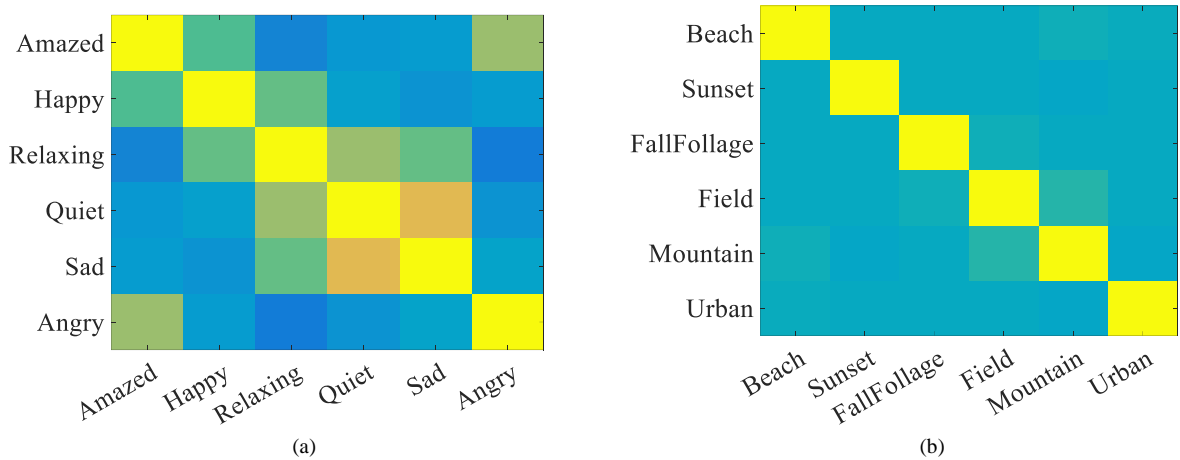

Fig. 4 Visualization of the correlation between any two labels (expressed by the true label matrix **Y**) on the (a) Emotions and (b) Scene datasets.

"Beach" and "Sunset" are likely to appear together, while it is unlikely for "Mountain" and "Urban" to be so.

The above analysis indicates that it is reasonable to use the correlation between $\tilde{\mathbf{y}}_i$ and $\tilde{\mathbf{y}}_j$ to train the correlation between $\mathbf{p}_{g,i}$ and $\mathbf{p}_{g,j}$. By this strategy, for similar labels, the related model parameters would be more correlated and have similar values, which is favorable for improving the performance of the multi-label classification task.

To further verify the effectiveness of label correlation learning in ML-TSK FS, two groups of experiments, i.e., Group A and Group B, are conducted in Fig. 5. The settings are as follows. For the parameter $\alpha$, which is used to adjust the influence of the label correlation learning term in (27), it is set to zero for Group A, and set to an appropriate value for Group B. That is, label correlation learning is disabled in Group A and enabled in Group B. The settings of the other parameters are the same for Group A and Group B.

It can be seen from the experiment results in Fig. 5 that the introduction of label correlation learning can indeed improve the classification performance of the ML-TSK FS.

### 3) Parameter Analysis

The performance of ML-TSK FS with respect to the parameters $K$, $\alpha$ and $\beta$ is studied in this section. The experiments and the results are discussed as follows.

For the number of rules $K$, we use the grid search strategy to optimize the number of rules for each dataset and the search range is {1,2,3,4,5,6,7,8,9,10}. The results on the 12 datasets, presented with three sub-figures for easy observation, are shown in Fig. 6. In general, ML-TSK FS can achieve optimal performance on the 12 datasets when the number of rules is between 2 and 6. If the number of rules is regarded as a hyperparameter, the optimal setting of this parameter can be determined automatically by grid search and cross-validation strategy.

The parameters $\alpha$ and $\beta$ are used respectively to adjust the weight of the correlation learning and the model complexity. Sensitivity analysis is conducted on the Flags dataset by first fixing $\beta$ while adjusting $\alpha$ to study the effect on AP. The analysis is then repeated by fixing $\alpha$ and adjusting $\beta$. To fully test the parameter sensitivity, $\alpha$ and $\beta$ are adjusted within a wide range of values in the set {0.01, 0.02, 0.03, …, 0.09, 0.1,

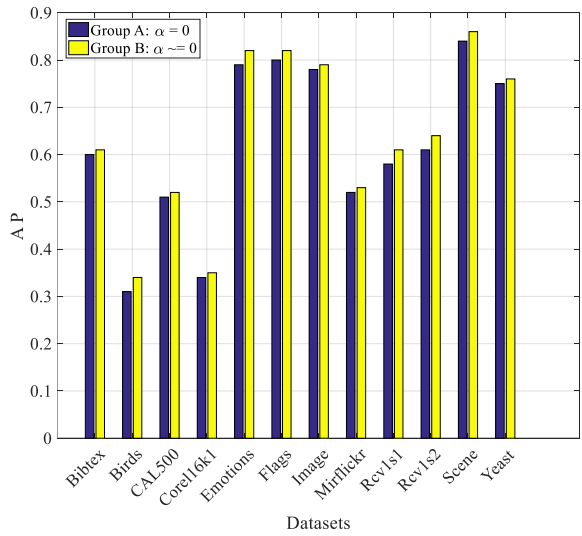
(a) AP

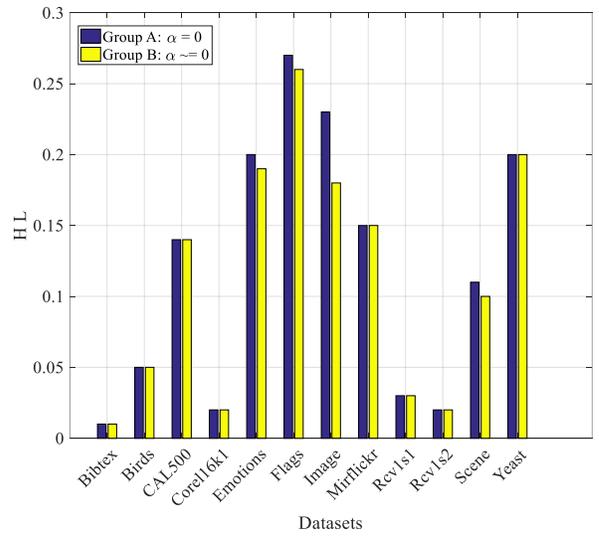
(b) HL

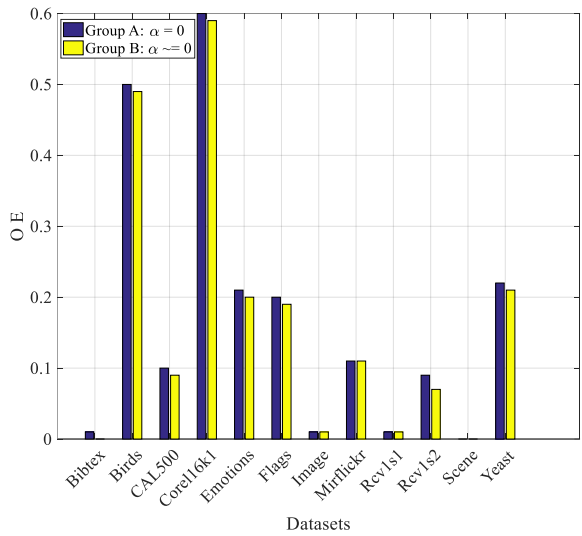
(c) OE

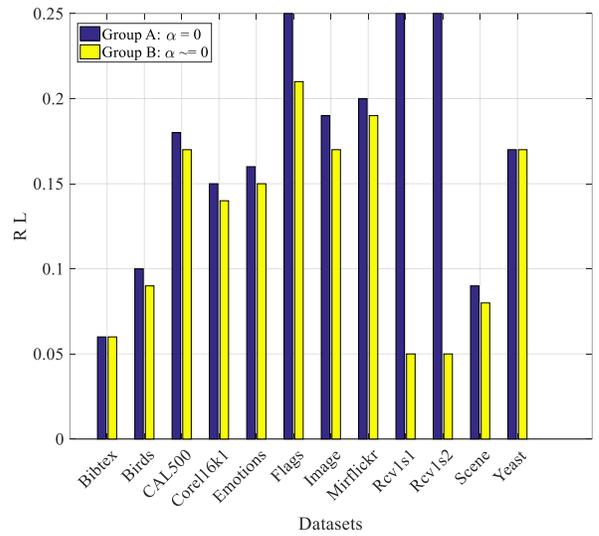
(d) RL

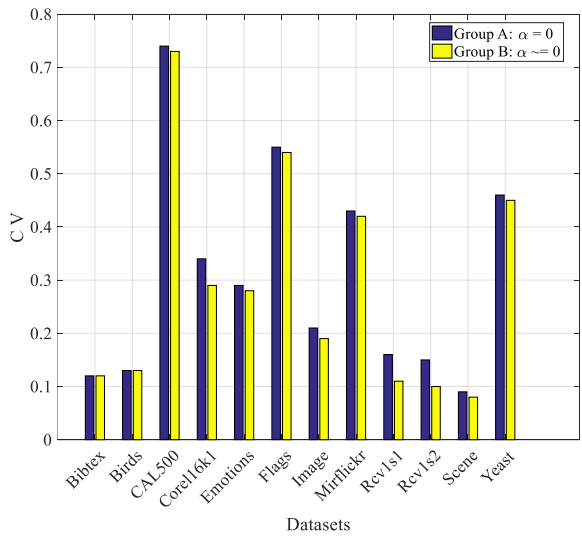
(e) CV

Fig. 5 Effect of correlation learning in ML-TSK FS in terms of five performance metrics. Correlation learning is disabled for the experiments in Group A, and is enabled for Group B. For AP, the larger the value, the better the classification performance; for the other metrics, the smaller the value, the better the classification performance.

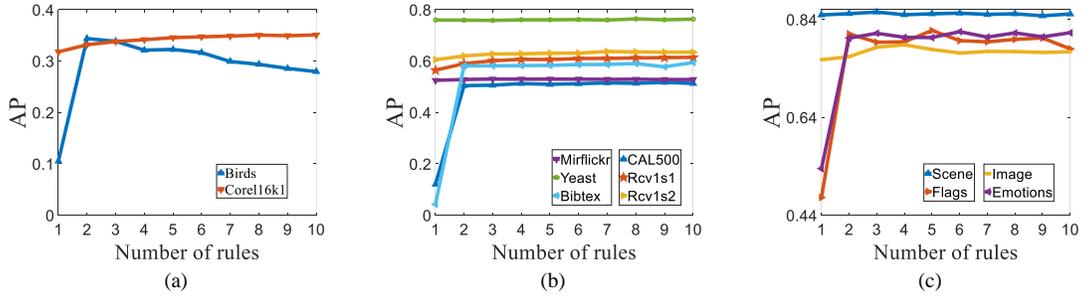

Fig. 6 Variation in AP with the numbers of rules, presented with three sub-figures for easy observation.

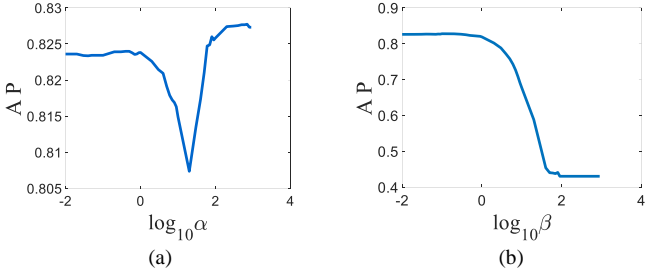

Fig. 7 Parameter sensitivity of (a) $\alpha$ and (b) $\beta$ on the Flags dataset.

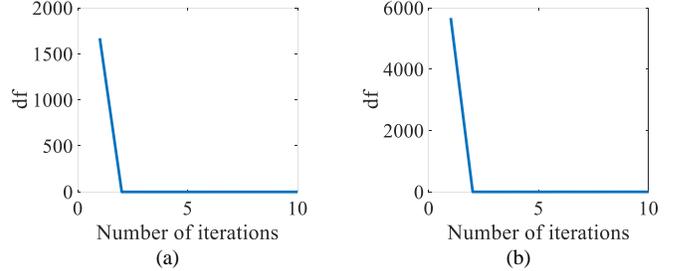

Fig. 8 Convergence analysis for the datasets (a) Scene and (b) Yeast.

0.2, 0.3, …, 0.9, 1, 2, 3, …, 9, 10, 20, 30, …, 90, 100, 200, 300, …, 900}. It can be seen from the results in Fig. 7 that the performance of ML-TSK FS is relatively stable when $\alpha$ is within [0.01, 0.6]. The performance fluctuates greatly when $\alpha$ is within [0.6, 60]. In the range [60, 900], AP increases with $\alpha$ and tends to the best values. On the other hand, the performance of ML-TSK FS is stable with high AP when $\beta$ is within [0.01, 0.9]. Beyond this range, AP decreases rapidly with increasing $\beta$.

*4) Convergence Analysis*

The convergence of ML-TSK FS is investigated experimentally by evaluating the absolute value of the difference between the previous and the current value of the objective function (denoted by df) as the algorithm iterates. The results of the experiments conducted using the datasets Scene and Yeast are shown in Fig. 8, which indicates that ML-TSK FS can converge within five iterations.

*5) Statistical Analysis*

Friedman test and Bonferroni-Dunn test [52] are used to analyze whether the performance difference between ML-TSK FS and the other methods are statistically significant. Based on the performance of the methods as shown in Table V, Friedman test is conducted on the five metrics AP, HL, OE, RL and CV. The null hypothesis is that there is no difference in performance between the methods in terms of the five metrics. If the Friedman statistic $F_F$ is greater than a critical value, the null hypothesis is rejected. Table VI shows that the null hypothesis is rejected for all the five metrics and the difference in performance observed are significant. The post-hoc Bonferroni-Dunn test is then conducted with ML-TSK FS being the reference method. The difference between the average rank of ML-TSK FS and another method is expressed in terms of critical difference (CD), which is given by

$$CD = q_\alpha \sqrt{k(k+1)/6M}$$

where $k$ and $M$ are the number of methods ($k = 9$) and the number of datasets ($M = 12$), respectively. With confidence level $\alpha = 0.05$ and $q_\alpha = 2.724$, we have CD = 3.0455. Fig. 9 shows the average ranks of the nine methods arranged from right (more superior) to left in ascending order of rank. The smaller the value of the rank, the more superior the method. If the difference in average rank between ML-TSK FS and a method is less than one CD, the difference in rank is considered not significant; otherwise, the difference is significant. In Fig. 9, ML-TSK FS is connected to a method with a red line if there is no significant difference in average rank between them. The following conclusions can be drawn from Fig. 9:

(1) ML-TSK FS is significantly superior to ML-$k$NN, CC, C2AE and JBNN in terms of AP, OE, RL and CV.

(2) Although the performance of ML-TSK FS is not significantly different from that of HNOML, BR, MLSF and BP-MLL, ML-TSK FS outperforms these four methods in general as evident from the results in Table V.

TABLE VI
FRIEDMAN STATISTICS ($k = 9$, $M = 12$)

| Evaluation metric | $F_F$ | Critical value ($\alpha$=0.05) |
|---|---|---|
| AP | 15.01 | |
| HL | 9.89 | |
| OE | 14.27 | 2.05 |
| RL | 17.64 | |
| CV | 8.03 | |

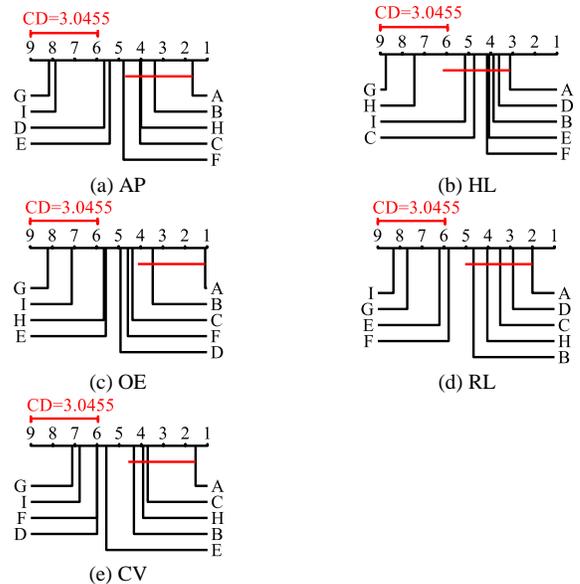

Fig. 9 Comparison of ML-TSK FS with other methods using the Bonferroni-Dunn test, in terms of (a) AP, (b) HL, (c) OE, (d) RL and (e) CV, with ML-TSK FS being the reference method. In the figures, the 9 methods are denoted by capital letters A to I, referring to ML-TSK FS, BR, HNOML, MLSF, ML-$k$NN, CC, C2AE, BP-MLL and JBNN respectively.

## V. Conclusion

In this paper, we propose the fuzzy multi-label classification method ML-TSK FS to improve the classification performance by introducing fuzzy inference and utilizing label correlation information. The ML-TSK FS has two distinctive characteristics. First, ML-TSK FS uses the rule structure to model the hidden association between features and labels. Second, ML-TSK FS utilizes the label correlation information to further enhance the classification abilities.

Although ML-TSK FS has shown promising performance, there is still room for improvement. One issue is that all the features are used by each of the fuzzy rules in the ML-TSK FS. If the dimension of the original feature space is large, ML-TSK FS will produce lengthy rules which increase the computational cost significantly and weaken the interpretability of the generated rules. Future work will be conducted to address the issue.

## Reference

[1] M. L. Zhang and Z. H. Zhou, "A review on multi-label learning algorithms," *IEEE Trans. Knowl. Data Eng.*, vol. 26, no. 8, pp. 1819-1837, Aug. 2014.

[2] O. Luaces, J. Díez, J. Barranquero, and J. J. D. Coz, "Binary relevance efficacy for multilabel classification," *Prog. Artif. Intell.*, vol. 1, no. 4, pp. 303-313, Dec. 2012.

[3] J. Read, B. Pfahringer, G. Holmes, and E. Frank, "Classifier chains for multi-label classification," *Mach. Learn.*, vol. 85, no. 3, pp. 333-359, Jun. 2011.

[4] M. L. Zhang, "Lift: Multi-label learning with label-specific features," *IEEE Trans. Pattern Anal. Mach. Intell.*, vol. 37, no. 1, pp. 107-120, Jan. 2015.

[5] J. A. Hartigan and M. A. Wong, "A k-means clustering algorithm," *Applied Stats.*, vol. 28, no. 1, pp. 100-108, Jan. 1979.

[6] J. Fürnkranz, E. Hüllermeier, E. L. Mencía, and K. Brinker, "Multilabel classification via calibrated label ranking," *Mach. Learn.*, vol. 73, no. 2, pp. 133-153, Nov. 2008.

[7] A. Clare and R. D. King, "Knowledge discovery in multi-label phenotype data," in *Proc. PKDD*, 2001, pp. 42-53.

[8] A. E. Elisseeff and J. Weston, "A kernel method for multi-labelled classification," in *Proc. Int. Conf. Neural Inf. Process. Syst.: Natural Synthetic*, 2001, pp. 681-687.

[9] M. L. Zhang, "LIFT: Multi-label learning with label-specific features," in *Proc. Int. Joint Conf. Artif. Intell.*, 2011, pp. 1609-1614.

[10] L. Hua, L. De-Yu, S. G. Wang, and J. Zhang, "Multi-label learning with label-specific features based on rough sets," *J. Chinese Comput. Syst.*, vol. 36, no. 12, pp. 2730-2734, Dec. 2015, doi: JournalArticle/5b3b6190c095d70f00708bd1.

[11] J. Chen, Y. Y. Tang, C. L. Chen, B. Fang, Y. Lin, and Z. Shang, "Multi-label learning with fuzzy hypergraph regularization for protein subcellular location prediction," *IEEE Trans. NanoBiosci.*, vol. 13, no. 4, pp. 438-447, Dec. 2014.

[12] M. L. Zhang and Z. H. Zhou, "ML-KNN: A lazy learning approach to multi-label learning," *Pattern Recognit.*, vol. 40, no. 7, pp. 2038-2048, Jul. 2007.

[13] W. Zhan and M. L. Zhang, "Multi-label learning with label-specific features via clustering ensemble," in *Proc. IEEE Int. Conf. Data Sci. Adv. Anal.*, 2017, pp. 129-136.

[14] S. M. García, C. J. Mantas, J. G. Castellano, and J. Abellán, "Ensemble of classifier chains and Credal C4.5 for solving multi-label classification," *Prog. Artif. Intell.*, vol. 8, no. 5, Jan. 2019.

[15] N. Li and Z. H. Zhou, "Selective ensemble of classifier chains," in *Proc. Int. Workshop on Mult. Class. Syst.*, 2013, pp. 146-156.

[16] Y. U. Zhang and D. Y. Yeung, "Multilabel relationship learning," *ACM Trans. Knowl. Discov. Data*, vol. 7, no. 2, pp. 1-30, Jan. 2013.

[17] S. Ji, L. Tang, S. Yu, and J. Ye, "Extracting shared subspace for multi-label classification," in *Proc. 14th ACM SIGKDD Int. Conf. Knowl. Discov. Data Min.*, 2008, pp. 381-389.

[18] Y. Guo and S. Gu, "Multi-label classification using conditional dependency networks," in *Proc. Int. Joint Conf. Artif. Intell.*, 2011, pp. 1300-1305.

[19] C. Li, B. Wang, V. Pavlu, and J. Aslam, "Conditional bernoulli mixtures for multi-label classification," in *Proc. Int. Conf. Mach. Learn.*, 2016, pp. 2482-2491.

[20] M. Biglarbegian, W. W. Melek, and J. M. Mendel, "On the stability of interval type-2 TSK fuzzy logic control systems," *IEEE Trans. Syst. Man. Cybern. B Cybern.*, vol. 40, no. 3, pp. 798-818, Jun. 2010.

[21] Q. F. Fan, T. Wang, Y. Chen, and Z. F. Zhang, "Design and application of interval type-2 TSK fuzzy logic system based on QPSO algorithm," *Int. J. Fuzzy Syst.*, vol. 20, no. 3, pp. 835-846, Jul. 2018.

[22] J. S. Guan, C. M. Lin, G. L. Ji, L. W. Qian, and Y. M. Zheng, "Robust adaptive tracking control for manipulators based on a TSK fuzzy cerebellar model articulation controller," *IEEE Access*, 2017, doi: 10.1109/ACCESS.2017.2779940.

[23] S. Nemet, D. Kukolj, G. Ostojić, S. Stankovski, and D. Jovanović, "Aggregation framework for TSK fuzzy and association rules: interpretability improvement on a traffic accidents case," *Appl. Intell.*, vol. 49, no. 11, pp. 3909-3922, Nov. 2019.

[24] C. S. Ouyang, W. J. Lee, and S. J. Lee, "A TSK-type neurofuzzy network approach to system modeling problems," *IEEE Trans. Syst. Man. Cybern. B Cybern.*, vol. 35, no. 4, pp. 751-767, Aug. 2005.

[25] J. Tavoosi, A. A. Suratgar, and M. B. Menhaj, "Stability analysis of recurrent type-2 TSK fuzzy systems with nonlinear consequent part," *Neural Comput. Appl.*, vol. 28, no. 1, pp. 47-56, Jan. 2017.

[26] T. Zhou, F. Chung, and S. Wang, "Deep TSK fuzzy classifier with stacked generalization and triply concise interpretability guarantee for large data," *IEEE Trans. Fuzzy Syst.*, vol. 25, no. 5, pp. 1207-1221, Oct. 2017.

[27] T. Zhou, H. Ishibuchi, and S. T. Wang, "Stacked blockwise combination of interpretable TSK fuzzy classifiers by negative correlation learning," *IEEE Trans. Fuzzy Syst.*, vol. 26, no. 6, pp. 3327-3341, Apr. 2018.

[28] Z. Deng, K. S. Choi, F. L. Chung, and S. Wang, "Scalable TSK fuzzy modeling for very large datasets using minimal-enclosing-ball approximation," *IEEE Trans. Fuzzy Syst.*, vol. 19, no. 2, pp. 210-226, Apr. 2011.

[29] T. Zhang, Z. Deng, D. Wu, and S. Wang, "Multiview fuzzy logic system with the cooperation between visible and hidden views," *IEEE Trans. Fuzzy Syst.*, vol. 27, no. 6, pp. 1162-1173, Jul. 2018.

[30] Y. Jiang, Z. Deng, F. L. Chung, G. Wang, P. Qian, K.-S. Choi, *et al.*, "Recognition of epileptic EEG signals using a novel multiview TSK fuzzy system," *IEEE Trans. Fuzzy Syst.*, vol. 25, no. 1, pp.

[31] Y. Jiang, Z. Deng, K. S. Choi, F. L. Chung, and S. Wang, "A novel multi-task TSK fuzzy classifier and its enhanced version for labeling-risk-aware multi-task classification," *Inf. Sci.*, vol. 357, no. 20, pp. 39-60, Aug. 2016.

[32] Z. Deng, P. Xu, L. Xie, K. S. Choi, and S. Wang, "Transductive joint-knowledge-transfer TSK FS for recognition of epileptic EEG signals," *IEEE Trans. Neural Syst. Rehabil. Eng.*, vol. 26, no. 8, pp. 1481-1494, Jun. 2018.

[33] C. Yang, Z. Deng, K. S. Choi, and S. Wang, "Takagi-Sugeno-Kang transfer learning fuzzy logic system for the adaptive recognition of epileptic electroencephalogram signals," *IEEE Trans. Fuzzy Syst.*, vol. 24, no. 5, pp. 1079-1094, Jan. 2015.

[34] S. S. Kim and K. C. Kwak, "Incremental modeling with rough and fine tuning method," *Appl. Soft. Comput.*, vol. 11, no. 1, pp. 585-591, Jan. 2011.

[35] W. Na and Y. Yang, "A fuzzy modeling method via enhanced objective cluster analysis for designing TSK model," *Expert Syst. Appl.*, vol. 36, no. 10, pp. 12375-12382, Dec. 2009.

[36] N. Kwak, "Principal component analysis based on l1-norm maximization," *IEEE Trans. Pattern Anal. Mach. Intell.*, vol. 30, no. 9, pp. 1672-1680, Sep. 2008.

[37] H. Wang, X. Lu, Z. Hu, and W. Zheng, "Fisher discriminant analysis with l1-norm," *IEEE Trans. Cybern.*, vol. 44, no. 6, pp. 828-842, Jun. 2014.

[38] F. Kahl, R. Hartley, and S. Member, "Multiple-view geometry under the l1-norm," *IEEE Trans. Pattern Anal. Mach. Intell.*, vol. 30, no. 9, pp. 1603-1617, Sep. 2008.

[39] A. Eriksson, V.D.H. Anton, "Efficient computation of robust weighted low-rank matrix approximations using the l1 norm," *IEEE Trans. Pattern Anal. Mach. Intell.*, vol. 34, no. 9, pp. 1681-1690, Sep. 2012.

[40] M. Carrasco, J. López, and S. Maldonado, "A multi-class SVM approach based on the l1-norm minimization of the distances between the reduced convex hulls," *Pattern Recognit.*, vol. 48, no. 5, pp. 1598-1607, May. 2015.

[41] P. C. Coefficient, "Pearson's correlation coefficient," *N. Z. Med. J.*, vol. 109, no. 1015, pp. 38-38, Feb. 1996.

[42] P. L. Combettes and V. R. Wajs, "Signal recovery by proximal forward-backward splitting," *Multiscale Model. Simul.*, vol. 4, no. 4, pp. 1168-1200, Jan. 2005.

[43] Z. Lin, A. Ganesh, J. Wright, L. Wu, M. Chen, and Y. Ma, "Fast convex optimization algorithms for exact recovery of a corrupted low-rank matrix," in *Proc. CAMSAP*, 2009, pp. 1-8.

[44] C. K. Yeh, W. C. Wu, W. J. Ko, and Y. C. F. Wang, "Learning deep latent spaces for multi-label classification", in *Proc. 31th AAAI Conf. Artif. Intell.*, 2017, pp. 2838-2844.

[45] C. Zhang, Z. Yu, H. Fu, P. Zhu, L. Chen, and Q. Hu, "Hybrid noise-oriented multilabel learning," *IEEE Trans. Cybern.*, vol. 50, no. 6, pp. 2837-2850, Jun. 2019.

[46] M. L. Zhang and Z. H. Zhou, "Multilabel neural networks with applications to functional genomics and text categorization," *IEEE Trans. Knowl. Data Eng.*, vol. 18, no. 10, pp. 1338-1351, Nov. 2006.

[47] H. He and R. Xia, "Joint binary neural network for multi-label learning with applications to emotion classification," in *Proc. CCF Int. Conf. Nat. Lang. Process. Chin. Comput.*, 2018, pp. 250-259.

[48] J. Łęski, "Improving the generalization ability of neuro-fuzzy systems by ε-insensitive learning," *Int. J. Appl. Math. Comput. Sci.*, vol. 12, no. 3, pp. 437-447, Jan. 2002.

[49] H. Li, D. Li, Y. Zhai, S. Wang, and J. Zhang, "A novel attribute reduction approach for multi-label data based on rough set theory," *Inf. Sci.*, vol. 367-368, no. 1, pp. 827-847, Nov. 2016.

[50] Y. Zhang and Z. H. Zhou, "Multi-label dimensionality reduction via dependence maximization," *ACM Trans. Knowl. Discov. Data*, vol. 4, no. 3, pp. 1-21, Oct. 2010.

[51] C. Li, C. Liu, L. Duan, P. Gao, and K. Zheng, "Reconstruction regularized deep metric learning for multi-label image classification," *IEEE Trans. Neural Netw. Learn. Syst.*, vol. 31, no. 7, pp. 2294-2303, Jul. 2020.

[52] O. J. Dunn, "Multiple comparisons among means," *J. Amer. Statist. Assoc.*, vol. 56, no. 293, pp. 52-64, Mar. 1961.

[53] J. Adler and I. Parmryd, "Quantifying colocalization by correlation: The Pearson correlation coefficient is superior to the Mander's overlap coefficient," *Cytom. Part A*, vol. 77a, no. 8, pp. 733-742, Aug. 2010.

[54] X. Zhang, "Matrix analysis and applications," Tsinghua, China: Tsinghua Univ. Press, 2004.

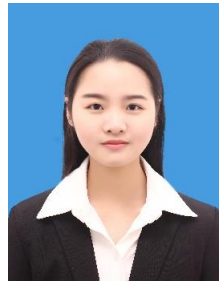

**Qiongdan Lou** received the B.S. degree in software engineering from Jiangsu University of Science and Technology, Suzhou, China, in 2017. She is currently pursuing the Ph.D. degree in the School of Artificial Intelligence and Computer Science, Jiangnan University, Wuxi, China.

Her research interests include interpretability and uncertainty artificial intelligence and pattern recognition.

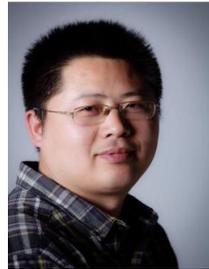

**Zhaohong Deng** (M'12-SM'14) received the B.S. degree in physics from Fuyang Normal College, Fuyang, China, in 2002, and the Ph.D. degree in information technology and engineering from Jiangnan University, Wuxi, China, in 2008.

He is currently a Professor with the School of Artificial Intelligence and Computer Science, Jiangnan University. He has visited the University of California-Davis and the Hong Kong Polytechnic University for more than two years. His current research interests include interpretability and uncertainty artificial intelligence and its applications. He has authored or coauthored more than 100 research papers in international/national journals.

Dr. Deng has served as an Associate Editor or Guest Editor of several international Journals, such as *IEEE TRANSACTIONS ON EMERGING TOPICS IN COMPUTATIONAL INTELLIGENCE*, *NEUROCOMPUTING*, and *PLOS ONE*.

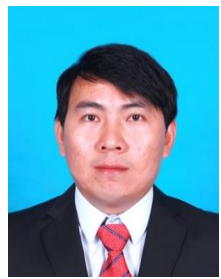

**Zhiyong Xiao** received the B.S. degree from Shandong University, China, in 2008, and the Ph.D. degree in Optics, Physics and Image Processing from the Ecole Central de Marseille, France, in 2014.

He is currently an Associate Professor with the School of Artificial Intelligence and Computer Science, Jiangnan University. He was an assistant research fellow at Fresnel Institute, CNRS France. His research interests include parallel computing, machine learning and artificial intelligence.

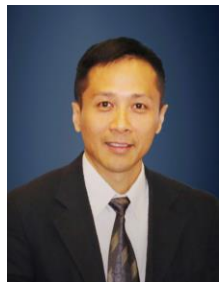

**Kup-Sze Choi** (M'97) received the Ph.D. degree in computer science and engineering from the Chinese University of Hong Kong, Hong Kong.

He is currently a Professor at the School of Nursing, Hong Kong Polytechnic University, Hong Kong, and the Director of the Centre for Smart Health. His research interests include virtual reality and artificial intelligence, and their applications in medicine and healthcare.

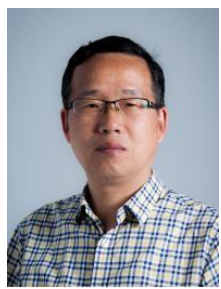

**Shitong Wang** received the M.S. degree in computer science from Nanjing University of Aeronautics and Astronautics, Nanjing, China, in 1987.

His research interests include artificial intelligence, neuro-fuzzy systems, pattern recognition, and image processing. He has published more than 100 papers in international/national journals and has authored 7 books.